\documentclass[10pt,journal,compsoc]{IEEEtran}
\usepackage{times}
\usepackage{epsfig}
\usepackage{graphicx}
\usepackage{amsmath}
\usepackage{amssymb}

\usepackage{multirow}
\usepackage{caption}
\usepackage{subcaption}
\usepackage{algorithm}
\usepackage{algpseudocode}
\usepackage{float}

\usepackage{color}
\usepackage{framed}
\usepackage{tabularx}
\usepackage{threeparttable}

\usepackage[pagebackref=false,breaklinks=true,letterpaper=true,colorlinks,bookmarks=false]{hyperref}

\def\eg{\emph{e.g. }} 
\def\ie{\emph{i.e. }}

\def\wrt{w.r.t. } 
\def\etal{\emph{et al. }}

\newcommand{\by} {{\bf y }}

\newcommand{\bk} {{\bf k }}

\newcommand{\bo} {{\bf o }}

\newcommand{\bI} {{\bf I }}

\newcommand{\vgg} {VGG$_{16}$}

\newcommand{\tabincell}[2]{\begin{tabular}{@{}#1@{}}#2\end{tabular}}

\begin{document}
%
\title{Deep Learning Markov Random Field \\ for Semantic Segmentation}
%
%
%
%

\author{Ziwei~Liu$^{*}$,
		Xiaoxiao~Li$^{*}$,
        Ping~Luo,\IEEEmembership{~Member,~IEEE},
        Chen~Change~Loy,\IEEEmembership{~Senior Member,~IEEE},
        and~Xiaoou~Tang,\IEEEmembership{~Fellow,~IEEE}
\IEEEcompsocitemizethanks{\IEEEcompsocthanksitem Z. Liu, X. Li, P. Luo, C. C. Loy and X. Tang are with the Department
of Information Engineering, The Chinese University of Hong Kong.\protect\\
E-mail: \{lz013, lx015, pluo, ccloy, xtang\}@ie.cuhk.edu.hk
\IEEEcompsocthanksitem * The first two authors share first-authorship. Correspondence to: Ping Luo and Ziwei Liu.}
}

%
%

\markboth{TO APPEAR IN IEEE TRANSACTIONS ON PATTERN ANALYSIS AND MACHINE INTELLIGENCE}%
{Shell \MakeLowercase{\textit{et al.}}: Bare Demo of IEEEtran.cls for Computer Society Journals}
%



\IEEEtitleabstractindextext{

\begin{abstract}

Semantic segmentation tasks can be well modeled by Markov Random Field (MRF). This paper addresses semantic segmentation by incorporating high-order relations and mixture of label contexts into MRF. Unlike previous works that optimized MRFs using iterative algorithm, we solve MRF by proposing a Convolutional Neural Network (CNN), namely Deep Parsing Network (DPN), which enables deterministic end-to-end computation in a single forward pass. Specifically, DPN extends a contemporary CNN to model unary terms and additional layers are devised to approximate the mean field (MF) algorithm for pairwise terms. It has several appealing properties. First, different from the recent works that required many iterations of MF during back-propagation, DPN is able to achieve high performance by approximating one iteration of MF. Second, DPN represents various types of pairwise terms, making many existing models as its special cases. Furthermore, pairwise terms in DPN provide a unified framework to encode rich contextual information in high-dimensional data, such as images and videos. Third, DPN makes MF easier to be parallelized and speeded up, thus enabling efficient inference. DPN is thoroughly evaluated on standard semantic image/video segmentation benchmarks, where a single DPN model yields state-of-the-art segmentation accuracies on PASCAL VOC 2012, Cityscapes dataset and CamVid dataset.

\end{abstract}


}

\maketitle

\IEEEdisplaynontitleabstractindextext

%
\IEEEpeerreviewmaketitle


%
%
%
%
\IEEEraisesectionheading{\section{Introduction}\label{sec:intro}}

\IEEEPARstart{S}{emantic} segmentation is a fundamental and long-standing problem in computer vision.
It is defined as a multi-label classification problem, aiming to assign each pixel with a category label.
%
{There are two widely adopted research realms, including semantic image segmentation \cite{shi2000normalized, felzenszwalb2006efficient, liu2011nonparametric, farabet2013learning, mostajabi2014feedforward, long2014fully, liu2015semantic} and semantic video segmentation \cite{gu1998semiautomatic, BrostowFC:PRL2008, lezama2011track, jain2013coarse, liu2015multiclass, tripathi2015semantic, kundu2016feature}. The former employs a static image as input, while the latter employs a video sequence.}
The obtained per-pixel segmentation results are extremely useful for several applications like smart editing \cite{isola2013scene}, scene understanding \cite{farabet2013learning} and automated driving \cite{Cordts2016Cityscapes}.

Since pixels in natural images or videos generally exhibit strong correlation, jointly modeling label distribution in all locations is desirable.
To capture these contextual information, Markov random field (MRF) and conditional random field (CRF) \cite{jordan1999introduction} are commonly used as classic frameworks for semantic segmentation.
They model the joint distribution of labels by defining both unary term and pairwise terms.
Unary term reflects the per-pixel confidence of assigning labels while pairwise terms capture the inter-pixel constraints.

Most previous studies focus on designing pairwise terms that possess strong expressive power.
For example, Kr{\"a}henb{\"u}hl \etal \cite{koltun2011efficient} attained accurate segmentation boundary by inferring on a fully-connected graph.
Vineet \etal \cite{vineet2012filter} extended \cite{koltun2011efficient} by defining both high-order and long-range terms between pixels.
Global or local semantic contexts between labels were also investigated by \cite{yang2014context}.
However, their performance are limited by the relatively shallow models (\eg SVM or Adaboost) used as unary term.
As deep learning gradually takes over in many image recognition fields \cite{simonyan2014very}, researchers have also explored the possibility of designing effective deep architecture for semantic segmentation.
For instance, Long \etal \cite{long2014fully} transformed fully-connected layers of CNN into convolutional layers, making accurate per-pixel classification possible using contemporary CNN architectures that were pre-trained on ImageNet \cite{deng2009imagenet}.
Chen \etal \cite{chen2014semantic} improved \cite{long2014fully} by feeding the outputs of CNN into a MRF with simple pairwise potentials,
but it treated CNN and MRF as separate components.
A recent advance was made in joint training CNN and MRF by passing the error of MRF inference backward into CNN~\cite{schwing2015fully}. Nonetheless, an iterative inference of MRF such as the mean field algorithm (MF) \cite{opper2001naive} is required for each training image during the back-propagation (BP).
Zheng \etal \cite{zheng2015conditional} further showed that the procedure of MF inference can be represented as a Recurrent Neural Network (RNN), but their computational costs are similar to that of~\cite{schwing2015fully}.

{We observed that a direct combination of CNN and MRF as above is inefficient,
posing challenges on both optimization difficulty and inference speed.
Since CNN typically has millions of parameters and MRF typically has thousands of latent variables,
they are cumbersome to jointly optimize and infer.}
Even worse, incorporating complex pairwise terms into a MRF becomes impractical,
limiting the performance of the entire system.
In this study, we propose a novel Deep Parsing Network (DPN),
which is an end-to-end system enabling jointly training of CNN and complex pairwise terms.
%
%
DPN has several appealing \textbf{properties}:

\vspace{0.1cm}
\noindent
(1) DPN solves MRF with a single feed-forward pass, reducing computational cost and meanwhile maintaining high performance.
Specifically, DPN models unary terms by extending the VGG-16 network (\vgg) \cite{simonyan2014very} pre-trained on ImageNet, while additional layers are carefully designed to model complex pairwise terms.
%
The learning of these terms is transformed into deterministic end-to-end computation by BP,
%
instead of embedding MF into BP as \cite{schwing2015fully, lin2015efficient} did.
Although MF can be represented by a RNN \cite{zheng2015conditional}, it needs to recurrently compute the forward pass so as to achieve good performance and thus the process is time-consuming,
\eg each forward pass contains hundred thousand of weights.
%
DPN approximates MF by using only one iteration of inference.
This is made possible by joint learning strong unary terms and rich pairwise information.
%
%

\vspace{0.1cm}
\noindent
(2) Pairwise terms determine the graphical structure.
In previous studies, if the former is changed, so is the latter as well as its inference procedure.
But with DPN, modifying the complexity of pairwise terms, \eg range of pixels and contexts, is as simple as modifying the receptive fields of convolutions, without varying BP.
Furthermore, DPN is capable of representing multiple types of pairwise terms, making many previous works \cite{chen2014semantic, zheng2015conditional, schwing2015fully} as its special cases.

\vspace{0.1cm}
\noindent
(3) 
DPN approximates MF with convolutional and pooling operations, which
can be speeded up by low-rank approximation \cite{jaderberg2014speeding} and easily parallelized \cite{chetlur2014cudnn} in a Graphical Processing Unit (GPU).

%

Our \textbf{contributions} are summarized as below. (1) We propose the novel DPN to jointly train \vgg~for unary terms with rich pairwise information, \ie \emph{mixture of label contexts} and \emph{high-order relations}.
In comparison to existing deep models, DPN approximates MF with only \emph{one iteration} of inference, reducing computational cost but still maintaining high performance.
(2) We show that multiple types of MRFs can be represented in DPN, making many previous works such as RNN \cite{zheng2015conditional} and DeepLab \cite{chen2014semantic} as its special cases.
(3) We conduct extensive experiments to investigate which component of DPN is crucial to achieve high performance.
%
%
%
We demonstrate the generalizability of DPN model by showing its state-of-the-art performance on several standard semantic image/video segmentation benchmarks, including PASCAL VOC 2012 \cite{everingham2010pascal}, CityScapes dataset \cite{Cordts2016Cityscapes} and CamVid dataset \cite{BrostowFC:PRL2008}.




In comparison to our earlier version of this work \cite{liu2015semantic}, we propose a generic deep learning framework, Deep Parsing Network (DPN) to model and solve $N$-Dimension ($N$-D) high-order Markov Random Field (MRF). Our previous study~\cite{liu2015semantic} only shows the possibility on 2-Dimension image segmentation problem.
Specifically, we employ dynamic node linking to construct graph in $N$-D space, which results in a model of $N$-D high-order MRF.
To solve this high-dimensional and high-order MRF, we re-formulate the mean field (MF) update process into a feed-forward pass of Convolutional Neural Network (CNN).
$N$-D local and global convolutional layers are designed to approximate different terms in a MF solver.
Apart from the methodology, the paper was also substantially improved by providing more technical details and more extensive experimental evaluations.

\section{Related Work}

Existing studies \cite{shi2000normalized, ren2003learning, felzenszwalb2006efficient, szummer2008learning, fulkerson2009class, arbelaez2011contour, farabet2013learning, mostajabi2014feedforward, long2014fully, luo2017learning, luo2017deep} on semantic segmentation focus on either constructing specific graph structure so that contextual information and long-term dependencies can be captured, or designing suitable network architecture to leverage the power of deep learning. In the following, we summarize recent research advances with respect to these two aspects. 

\vspace{0.1cm} \noindent \textbf{Markov Random Field.~}
Markov Random Field (MRF) or Conditional Random Field (CRF) has achieved great successes in semantic image segmentation, which is one of the most challenging problems in computer vision.
%
%
Researchers improved labeling accuracy by exploring rich information to define the pairwise functions, including long-range dependencies \cite{koltun2011efficient, krahenbuhl2013parameter}, high-order potentials \cite{vineet2012filter, vineet2013posefield}, and semantic label contexts \cite{liu2011nonparametric, mottaghi2014role, yang2014context}.
For example, Kr{\"a}henb{\"u}hl \etal \cite{koltun2011efficient} attained accurate segmentation boundary by inferring on a fully-connected graph.
Vineet \etal \cite{vineet2012filter} extended \cite{koltun2011efficient} by defining both high-order and long-range terms between pixels.
Global or local semantic contexts between labels were also investigated by \cite{yang2014context}.
Although they accomplished promising results,
they modeled the unary terms as SVM or Adaboost, whose learning capacity becomes a bottleneck.
The learning and inference of complex pairwise terms are often expensive.

MRF and CRF have also been utilized in semantic video segmentation by extending their graph structure to spatio-temporal domain.
For example, Wang \etal \cite{wang2009segmentation} unified foreground object segmentation, tracking and occlusion reasoning into a carefully designed MRF model.
Optical flow based long-term trajectories \cite{lezama2011track} were also exploited to discover moving objects.
Liu \etal \cite{liu2015multiclass} employed fully-connected CRF augmented with object potentials for efficient multi-class inference. 
However, these methods are based on hand-crafted features, thus lacking sufficient learning capacity.

\vspace{0.1cm}\noindent\textbf{Convolutional Neural Network.~}
More recently, Convolutional Neural Network (CNN) has been leveraged as a strong unary classifier.
With deep models, existing works \cite{luo2012hierarchical, luo2013pedestrian, mostajabi2014feedforward, long2014fully, chen2014semantic, papandreou2015weakly, zheng2015conditional, schwing2015fully, lin2015efficient} demonstrated encouraging segmentation results through using just simple definition of the pairwise function or even neglecting it.
%
For instance, Long \etal \cite{long2014fully} transformed fully-connected layers of CNN into convolutional layers, making accurate per-pixel classification possible using the contemporary CNN architectures that were pre-trained on ImageNet \cite{deng2009imagenet}.
Chen \etal \cite{chen2014semantic} improved \cite{long2014fully} by feeding the outputs of CNN into a MRF with simple pairwise potentials,
but it treated CNN and MRF as separated components.
A recent advance was obtained by \cite{schwing2015fully}, which jointly trained CNN and MRF by passing the error of MRF inference backward into CNN,
but iterative inference of MRF such as the mean field algorithm (MF) \cite{opper2001naive} is required for each training image during the back-propagation (BP).
Zheng \etal \cite{zheng2015conditional} further showed that the procedure of MF inference can be represented as a Recurrent Neural Network (RNN), but their computational costs are similar to that of~\cite{schwing2015fully}.
%
%

Little attempts have been made to develop unified deep learning framework for semantic video segmentation.
Recent efforts in this direction include SegNet \cite{badrinarayanan2015segnet}, which adopted an encoder-decoder architecture but did not take temporal relationships into consideration.
Here we extend DPN to further include temporal voxels into the joint learning and inference process, which results in an end-to-end trainable system with rich spatio-temporal information encoded.

\begin{figure*}[t]
  \centering
  \includegraphics[width=0.83\textwidth]{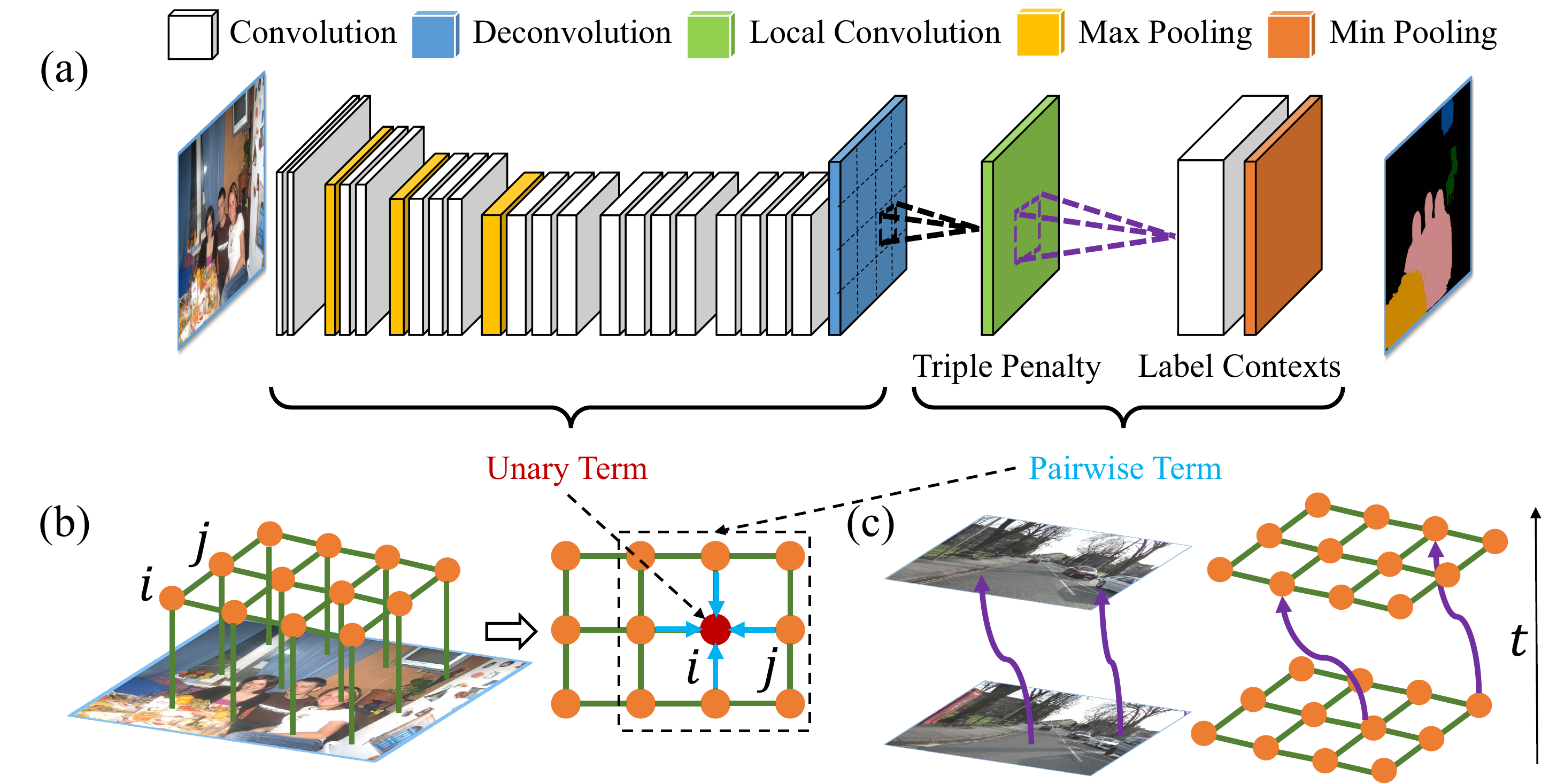}
  \caption{(a) The network architecture of a deep parsing network (DPN). (b) DPN extends a contemporary CNN architecture to model unary terms and additional layers are carefully devised to approximate the mean field algorithm (MF) for pairwise terms. (c) DPN enables dynamic linking of nodes in Markov Random Field (MRF) by incorporating domain knowledge.}
  \label{fig:pipeline}
  \vspace{-8pt}
\end{figure*}

\section{Our Approach}

We develop a unified framework, DPN, for modeling and solving high-order MRF. The architecture of DPN is shown in Fig.~\ref{fig:pipeline}(a).
Our model imposes no restrictions on the dimension of MRF.
For example, it can be either $2$-Dimension (2-D) for images segmentation, $3$-Dimension (3-D) for video segmentation, or $N$-Dimension (N-D) for sensor data.
DPN learns MRF by extending \vgg~to model unary terms and additional layers are carefully designed to model pairwise terms.
In the following, we describe the formulation of DPN in terms of $3$-D MRF.
Note that it can be easily resorted to $2$-D cases or extended to $N$-D cases by removing or adding relationships between nodes.

\vspace{4pt}\hspace{-15pt}\textbf{Markov Random Field.~} MRF \cite{freeman2000learning} is an undirected graph where each node represents a voxel in a video, $\bI$, and each edge represents relation between voxels, as shown in Fig.~\ref{fig:pipeline}(b).
Each node is associated with a binary latent variable, $y^u_{\mathbf{i}}\in\{0,1\}$, indicating whether a voxel $\mathbf{i} = [i~t_{i}]$ has label $u$.
Here, $i$ indicates a voxel's spatial index with respect to an image, and $t_{i}$ is its temporal index with respect to a sequence.
We have $\forall u\in L=\{1,2,...,l\}$, representing a set of $l$ labels.
The energy function of MRF is written as
\begin{equation}\label{eq:E}
E(\by)=\sum_{\forall \mathbf{i}\in\mathcal{V}}\Phi(y_{\mathbf{i}}^u)+\sum_{\forall (\mathbf{i}, \mathbf{j})\in\mathcal{E}}\Psi(y^u_{\mathbf{i}},y^v_{\mathbf{j}}),
\end{equation}
where $\by$, $\mathcal{V}$, and $\mathcal{E}$ denote a set of latent variables, nodes, and edges, respectively.
$\Phi(y^u_{\mathbf{i}})$ is the unary term, measuring the cost of assigning label $u$ to voxel $\mathbf{i}$.
For instance, if voxel $\mathbf{i}$ belongs to the first category other than the second one at time $t_i$, we should have $\Phi(y^1_{\mathbf{i}})<\Phi(y^2_{\mathbf{i}})$.
Moreover, $\Psi(y_{\mathbf{i}}^u,y_{\mathbf{j}}^v)$ is the pairwise term that measures the penalty of assigning labels $u,v$ to a pair of voxel $(\mathbf{i}, \mathbf{j})$ respectively.
%
%

\vspace{4pt}\hspace{-15pt}\textbf{Dynamic Node Linking.~}
Traditional approaches \cite{tripathi2015semantic, liu2014fast, liu2017video} usually define the edges $\mathcal{E}$ on rectangular grid in $3$-D space.
However, when large motion exists, the actual temporal trajectory for certain pixel will not reside inside a rigid cube, which means the rectangular grid assumption does not hold.
To better preserve the contextual information in a spatio-temporal space, we employ dynamic node linking to construct edges $\mathcal{E}$ in DPN.
Specifically, we keep the 2-D structure in the spatial domain, as illustrated in Fig.~\ref{fig:pipeline}(c).
In the temporal domain, the neighboring voxels $\mathbf{i}=[i~t_{i}]$ and $\mathbf{j}=[j~t_{j}]$ are defined as those lie on the same temporal trajectories $\Delta_{\mathbf{i} \rightarrow \mathbf{j}}$.
This trajectory can be estimated by standard optical flow techniques \cite{liu2011sift}.
The formulation of edges $\mathcal{E}_{t}$ in the temporal domain is
\begin{equation}\label{eq:edge}
(\mathbf{i}, \mathbf{j}) \in \mathcal{E}_{t} \iff \mathbf{j} = \mathbf{i} + \Delta_{\mathbf{i} \rightarrow \mathbf{j}}.
\end{equation}
In this setting, adjacent nodes in the temporal space would be more likely to belong to the same category, making the label contexts more easily to be captured.

\vspace{4pt}\hspace{-15pt}\textbf{Unary and Pairwise Terms.~}
Intuitively, the unary terms represent per-voxel classifications, while the pairwise terms represent a set of smoothness constraints.
The unary term in Eqn.~(\ref{eq:E}) is typically defined as
\begin{equation}\label{eq:unary}
\Phi(y_{\mathbf{i}}^u)=-\ln p(y_{\mathbf{i}}^u=1|\bI),
\end{equation}
where $p(y_{\mathbf{i}}^u=1|\bI)$ indicates the probability of the presence of label $u$ at voxel $\mathbf{i}$, modeling by \vgg.
To simplify discussions, we abbreviate it as $p^u_{\mathbf{i}}$.
%
%
The smoothness term can be formulated as
\begin{equation}\label{eq:old_p}
\Psi(y_{\mathbf{i}}^u,y_{\mathbf{j}}^v)=\mu(u,v)\mathrm{d}(\mathbf{i}, \mathbf{j}),
\end{equation}
where
the first term learns the penalty of global co-occurrence between any pair of labels. For example, the output value of $\mu(u,v)$ is large if $u$ and $v$ should not coexist.
In Eqn.(\ref{eq:old_p}), the second term calculates the distances between voxels. We have
\begin{equation}
\mathrm{d}(\mathbf{i},\mathbf{j})=\mathrm{d}(i,t_{i},j,t_{j})=\omega_1\|\bI_{\mathbf{i}}-\bI_{\mathbf{j}}\|^2+\omega_2\|[i~t_i]-[j~t_j]\|^2,
\end{equation}
where $\bI_{\mathbf{i}}$ indicates a feature vector such as RGB values extracted from the input video for voxel $\mathbf{i}$, $[i~t_i]$ denote coordinates of voxels' positions, and $\omega_1,\omega_2$ are the constant weights.
Eqn.~(\ref{eq:old_p}) implies that if two voxels are close and look similar, they are encouraged to have labels that are compatible.
This formulation has been adopted by most of the recent deep models \cite{chen2014semantic, zheng2015conditional, schwing2015fully} for semantic image segmentation.

However, Eqn.~(\ref{eq:old_p}) has two main drawbacks.
First, its first term captures the co-occurrence frequency of two labels in the training data, but neglects the spatial context between objects.
For example, `person' may appear beside `table', but not at its bottom.
This spatial context is a mixture of patterns, as different object configurations may appear in different images, such as `person' standing beside `table' and `person' sitting behind `table'.
Second, it defines only the pairwise relations between pixels, missing their high-order interactions.

\begin{figure}[t]
  \centering
  \includegraphics[width=0.45\textwidth]{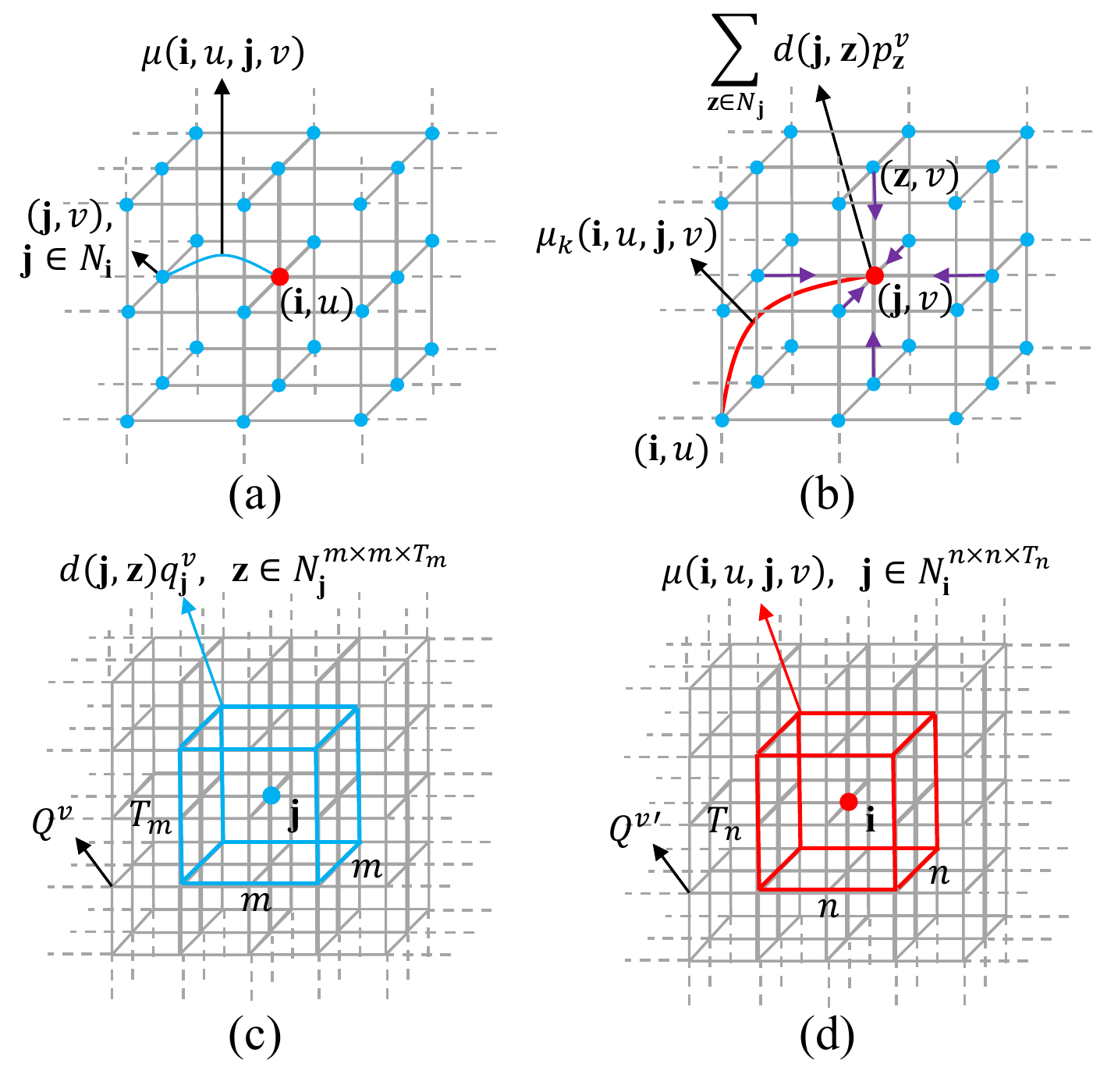}
  \vskip -0.2cm
  \caption{(a) Illustration of the pairwise terms in DPN. (b) explains the label contexts. (c) and (d) show that the mean field update of DPN corresponds to convolutions.}
  \label{fig:mrf}
  \vspace{-8pt}
\end{figure}

To resolve these issues, we define the smoothness term by leveraging rich information between voxels, which is one of the \textbf{advantages} of DPN over existing deep models.
%
We have
\begin{eqnarray}\label{eq:pairwise}
\Psi(y_{\mathbf{i}}^u,y_{\mathbf{j}}^v)&=&\sum_{k=1}^K\lambda_k\mu_k(\mathbf{i},u,\mathbf{j},v)\sum_{\forall \mathbf{z}\in \mathcal{N}_{\mathbf{j}}}\mathrm{d}(\mathbf{j}, \mathbf{z})p^v_{\mathbf{z}}.
\end{eqnarray}
The first term in Eqn.~(\ref{eq:pairwise})
%
learns a \textbf{mixture of local label contexts}, penalizing label assignment in a local cube,
where $K$ is the number of components in the mixture and $\lambda_k$ is an indicator, determining which component is activated. We define $\lambda_k\in\{0,1\}$ and $\sum_{k=1}^K\lambda_k=1$.
%
An intuitive illustration is given in Fig.~\ref{fig:mrf}(a),
where the dots in red and blue represent a center voxel $\mathbf{i}$ and its neighboring voxels $\mathbf{j}$, \ie $\mathbf{j}\in\mathcal{N}_{\mathbf{i}}$, and $(\mathbf{i},u)$ indicates assigning label $u$ to voxel $\mathbf{i}$.
Here, $\mu(\mathbf{i},u,\mathbf{j},v)$ outputs labeling cost between $(\mathbf{i},u)$ and $(\mathbf{j},v)$ with respect to their relative positions.
For instance, if $u,v$ represent `person' and `table', the learned penalties of positions $\mathbf{j}$ that are at the bottom of voxel $\mathbf{i}$ should be large.
%
%
The second term of Eqn.~(\ref{eq:pairwise}) basically models a \textbf{triple penalty}, which involves voxels $\mathbf{i}$, $\mathbf{j}$, and $\mathbf{j}$'s neighbors,
implying that if $(\mathbf{i},u)$ and $(\mathbf{j},v)$ are compatible, then $(\mathbf{i},u)$ should be also compatible with $\mathbf{j}$'s nearby pixels $(\mathbf{z},v)$, $\forall \mathbf{z}\in\mathcal{N}_{\mathbf{j}}$, as shown by the purple arrows in Fig.~\ref{fig:mrf}(b).

Learning parameters (\ie weights of \vgg~and costs of label contexts) in Eqn.~(\ref{eq:E}) requires us to
%
minimize the distances between ground-truth label map and the predicted label $\by$, which needs to be inferred subject to the smoothness constraints.

\vspace{4pt}\hspace{-15pt}\textbf{Inference Overview.~}
Inference of Eqn.~(\ref{eq:E}) can be obtained by the mean field (MF) algorithm \cite{opper2001naive}, which
estimates the joint distribution of MRF
\begin{equation}
P(\by) = \frac{1}{Z}\exp\{-E(\by)\},
\end{equation}
by using a fully-factorized proposal distribution
\begin{equation}
Q(\by) = \prod_{\forall \mathbf{i}\in\mathcal{V}}\prod_{\forall u\in L}q^u_{\mathbf{i}},
\end{equation}
where each $q_{\mathbf{i}}^u$ is a variable we need to estimate, indicating the predicted probability of assigning label $u$ to voxel $\mathbf{i}$.
The Kullback-Leibler divergence between them is then calculated as
\begin{equation}
\label{eqn:kl}
\begin{split}
D_{KL}(Q\|P) &= \sum_{\by} Q(\by) \ln \left( \frac{Q(\by)}{P(\by)} \right) \\
&= \sum_{\by} Q(\by)E(\by) + \sum_{\by} Q(\by) \ln Q(\by) + \ln Z.
\end{split}
\end{equation}
Since $\ln Z$ is a constant, minimizing the Kullback-Leibler divergence between $Q(\by)$ and $P(\by)$ is equivalent to minimizing the former terms in Eqn.~(\ref{eqn:kl}),
which is denoted as free energy $F(Q)$ \cite{jordan1999introduction}.
To simplify the discussion, we denote $\Phi(y_{\mathbf{i}}^u)$ and $\Psi(y_{\mathbf{i}}^u,y_{\mathbf{j}}^v)$ as $\Phi_{\mathbf{i}}^u$ and $\Psi_{\mathbf{ij}}^{uv}$, respectively.
And we can further substitute Eqn.(\ref{eq:E}) into $F(Q)$. We have
\begin{equation}\label{eq:FQ}
\begin{split}
F(Q) &= \sum_{\by} Q(\by)E(\by) + \sum_{\by} Q(\by) \ln Q(\by) \\
&= \sum_{\forall \mathbf{i}\in\mathcal{V}}\sum_{\forall u\in L}q_{\mathbf{i}}^u\Phi_{\mathbf{i}}^u+\sum_{\forall \mathbf{i,j}\in\mathcal{E}}\sum_{\forall u\in L}\sum_{\forall v\in L}q_{\mathbf{i}}^u q_{\mathbf{j}}^v \Psi_{\mathbf{ij}}^{uv} \\
&~~~~+ \sum_{\forall \mathbf{i}\in\mathcal{V}}\sum_{\forall u\in L} q_{\mathbf{i}}^u\ln q_{\mathbf{i}}^u.
\end{split}
\end{equation}
Specifically, the first term in Eqn.~(\ref{eq:FQ}) characterizes the cost of each voxel's predictions, while the second term characterizes the consistencies of predictions between voxels. The last term denotes the entropy, measuring the confidences of predictions.
Then, a constrained optimization problem regarding $q_{\mathbf{i}}^u$ could be formulated as
\begin{equation}
\begin{aligned}
& \underset{q_{\mathbf{i}}^u}{\text{minimize}}
& & F(Q) \\
& \text{subject to}
& & \sum_{u} q_{\mathbf{i}}^u = 1, \; \forall \mathbf{i}\in\mathcal{V}.
\end{aligned}
\end{equation}
To solve this minimization problem, we define $J(Q) = F(Q) + \sum_{\mathbf{i}}\lambda_{\mathbf{i}}(\sum_{u} q_{\mathbf{i}}^u - 1)$ by introducing Lagrange multipliers $\lambda_{\mathbf{i}}$.
The final closed-form solution can be obtained by differentiating $J(Q)$ \wrt to $q_{\mathbf{i}}^u$ and equating the resulting expression to zero
\begin{equation}\label{eq:q}
q_{\mathbf{i}}^u\propto\exp\big\{-(\Phi_{\mathbf{i}}^u+\sum_{\forall \mathbf{j}\in\mathcal{N}_{\mathbf{i}}}\sum_{\forall v\in L}q_{\mathbf{j}}^v\Psi_{\mathbf{ij}}^{uv})\big\},
\end{equation}
such that the predictions for each voxel is independently attained by repeatedly calculating Eqn.~(\ref{eq:q}),
which implies whether voxel $\mathbf{i}$ have label $u$ is proportional to the estimated probabilities of all its neighboring voxels, weighted by their corresponding smoothness penalties.
Substituting Eqn.~(\ref{eq:pairwise}) into Eqn.~(\ref{eq:q}), we have
\begin{eqnarray}\label{eq:newq}
q_{\mathbf{i}}^u&\propto& \exp\Big\{
-\Phi_{\mathbf{i}}^u-\sum_{k=1}^K\lambda_k\sum_{\forall v\in L}\\\nonumber
&&\sum_{\forall \mathbf{j}\in\mathcal{N}_{\mathbf{i}}}\mu_k(\mathbf{i},u,\mathbf{j},v)\sum_{\forall \mathbf{z}\in\mathcal{N}_{\mathbf{j}}}\mathrm{d}(\mathbf{j},\mathbf{z})q_{\mathbf{j}}^v q^v_{\mathbf{z}}
\Big\},
\end{eqnarray}
where each $q^u_{\mathbf{i}}$ is initialized by the corresponding $p^u_{\mathbf{i}}$ in Eqn.~(\ref{eq:unary}), which is the unary prediction of \vgg.
Eqn.~(\ref{eq:newq}) satisfies the smoothness constraints.

\begin{figure*}[t]
  \centering
  \includegraphics[width=0.9\textwidth]{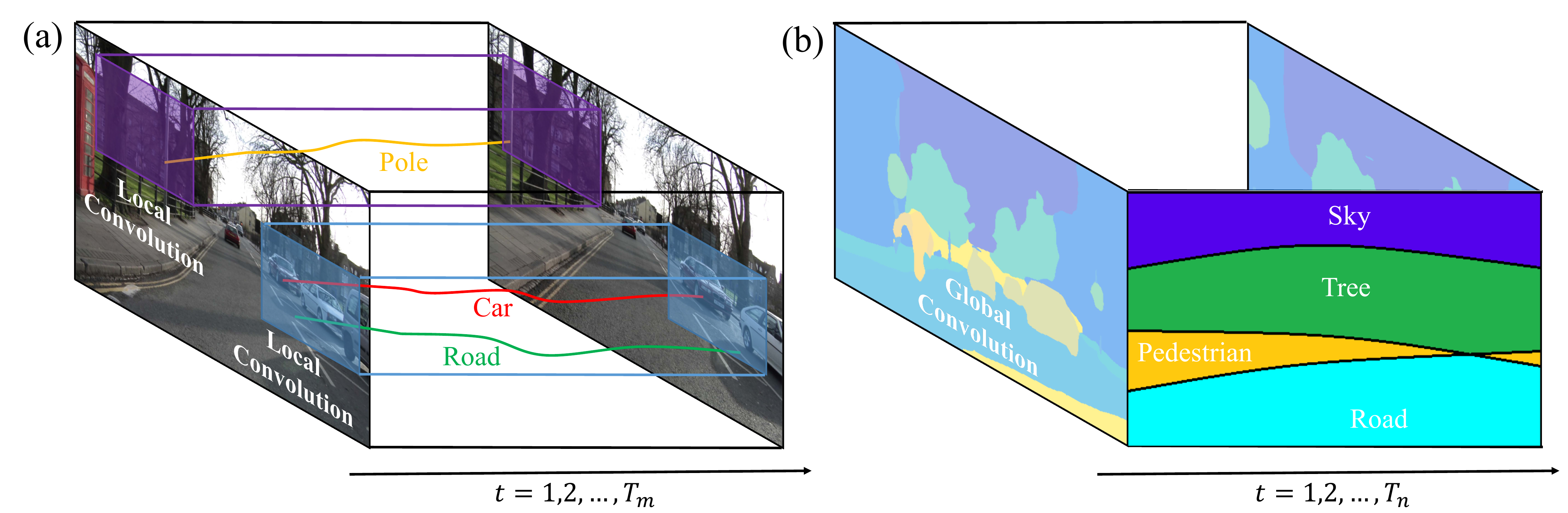}
  \caption{Illustrative depiction of (a) triple penalty term and (b) mixture of local label contexts term.}
  \label{fig:conv_3d}
  \vspace{-8pt}
\end{figure*}

In the following, DPN approximates one iteration of Eqn.~(\ref{eq:newq}) by decomposing it into two steps.
%
Let $Q^v$ be a predicted label map of the $v$-th category.
In the first step as shown in Fig.~\ref{fig:mrf}(c),
%
we calculate the triple penalty term in Eqn.~(\ref{eq:newq}) by applying a $m\times m \times T_{m}$ filter on each position $\mathbf{j}$, where
each element of this filter equals $\mathrm{d}(\mathbf{j},\mathbf{z})q^v_{\mathbf{j}}$, resulting in ${Q^v}'$.
$T_m$ indicates the time span.
%
%
Apparently, this step smoothes the prediction of voxel $\mathbf{j}$ with respect to the distances between it and its neighborhood.
In the second step as illustrated in Fig.~\ref{fig:mrf}(d),
the labeling contexts can be obtained by convolving ${Q^v}'$ with a $n\times n \times T_{n}$ filter, each element of which equals $\mu_k(\mathbf{i},u,\mathbf{j},v)$, penalizing the triple relations as shown in Fig.~\ref{fig:mrf}(a).

Fig.~\ref{fig:conv_3d} depicts the semantic meaning of the \textbf{triple penalty} and the \textbf{mixture of local label contexts} term in the spatial-temporal domain.
From Fig.~\ref{fig:conv_3d}(a) we can see that the triple penalty term tracks the movement of local pixels, such as `pole', `car' and `road'.
These temporal trajectories combined with local regions are subsequently used to smooth original predictions.
Fig.~\ref{fig:conv_3d}(b) demonstrates that a mixture of local label contexts term not only captures label co-occurrence in a single image, but also encodes the change of label configurations along time.
For example, as the observer vehicle drives, `tree' will move backward and get nearer to `road'.

\section{Deep Parsing Network}

\begin{table*}[t]
\scriptsize
\vspace{14pt}
\caption{A comparison between the network architectures of \vgg~and DPN.}
\label{tab:net}
\vspace{-14pt}
\begin{center}
\begin{tabular}{c|c|c|c|c|c|c|c|c|c|c|c|c|c|c|c}
\multicolumn{16}{c}{(a)~\textbf{\vgg:}~~224$\times$224$\times$3 \emph{input image};~~1$\times$1000 \emph{output labels}} \\
\hline
\multicolumn{1}{c}{} & \multicolumn{1}{c}{1} & \multicolumn{1}{c}{2} & \multicolumn{1}{c}{3} & \multicolumn{1}{c}{4} & \multicolumn{1}{c}{5} & \multicolumn{1}{c}{6} & \multicolumn{1}{c}{7} & \multicolumn{1}{c}{8} & \multicolumn{1}{c}{9} & \multicolumn{1}{c}{10} & \multicolumn{1}{c}{11} &
\multicolumn{1}{c}{12} &\multicolumn{1}{c}{} & \multicolumn{1}{c}{} & \multicolumn{1}{c}{}\\
\hline
\tabincell{r}{\emph{\textbf{layer}}\\\emph{\textbf{fi.-st.}}\\\emph{\textbf{\#ch.}}\\\emph{\textbf{act.}}\\\emph{\textbf{size}}} &
\tabincell{c}{2$\times$\textbf{conv}\\3-1\\64\\ $\mathrm{relu}$\\ 224} & 
\tabincell{c}{\textbf{max}\\2-2\\64\\ $\mathrm{idn}$\\112} &
\tabincell{c}{2$\times$\textbf{conv}\\3-1\\128\\$\mathrm{relu}$\\ 112} &
\tabincell{c}{\textbf{max}\\2-2\\128\\ $\mathrm{idn}$\\56} &
\tabincell{c}{3$\times$\textbf{conv}\\3-1\\256 \\$\mathrm{relu}$\\ 56} &
\tabincell{c}{\textbf{max}\\2-2\\256\\ $\mathrm{idn}$\\28} &
\tabincell{c}{3$\times$\textbf{conv}\\3-1\\ 512 \\$\mathrm{relu}$\\ 28} &
\tabincell{c}{\textbf{max}\\2-2\\512\\ $\mathrm{idn}$\\14} &
\tabincell{c}{3$\times$\textbf{conv}\\3-1\\ 512 \\$\mathrm{relu}$\\ 14} &
\tabincell{c}{\textbf{max}\\2-2\\512\\ $\mathrm{idn}$\\7} &
\tabincell{c}{2$\times$\textbf{fc}\\-\\1\\ $\mathrm{relu}$\\4096} &
\tabincell{c}{\textbf{fc}\\-\\1\\ $\mathrm{soft}$\\1000} &
\multicolumn{1}{c}{} &
\multicolumn{1}{c}{} & \multicolumn{1}{c}{}
\\
\hline
\multicolumn{16}{c}{} \\
\multicolumn{16}{c}{(b)~\textbf{DPN:}~~$T\times$512$\times$512$\times$3 \emph{input image};~~$T\times$512$\times$512$\times L$ \emph{output label maps}} \\
\hline
\multicolumn{1}{c}{} & \multicolumn{1}{c}{1} & \multicolumn{1}{c}{2} & \multicolumn{1}{c}{3} & \multicolumn{1}{c}{4} & \multicolumn{1}{c}{5} & \multicolumn{1}{c}{6} & \multicolumn{1}{c}{7} & \multicolumn{1}{c}{8} & \multicolumn{1}{c}{9} & \multicolumn{1}{c}{10} & \multicolumn{1}{c}{11} &
\multicolumn{1}{c}{12} & \multicolumn{1}{c}{13} & \multicolumn{1}{c}{14} &
\multicolumn{1}{c}{15} \\
\hline
\tabincell{r}{\emph{\textbf{layer}}\\\emph{\textbf{fi.-st.}}\\\emph{\textbf{\#ch.}}\\\emph{\textbf{act.}}\\\emph{\textbf{size}}} &
\tabincell{c}{2$\times$\textbf{conv}\\3-1\\64\\$\mathrm{relu}$\\$T$-512} &
\tabincell{c}{\textbf{max}\\2-2\\64\\$\mathrm{idn}$\\ $T$-256} &
\tabincell{c}{2$\times$\textbf{conv}\\3-1\\128\\ $\mathrm{relu}$\\$T$-256} &
\tabincell{c}{\textbf{max}\\2-2\\128\\ $\mathrm{idn}$\\$T$-128} &
\tabincell{c}{3$\times$\textbf{conv}\\3-1\\256 \\$\mathrm{relu}$\\$T$-128} &
\tabincell{c}{\textbf{max}\\2-2\\ 256\\ $\mathrm{idn}$\\$T$-64} &
\tabincell{c}{3$\times$\textbf{conv}\\3-1\\ 512 \\ $\mathrm{relu}$\\$T$-64} &
\tabincell{c}{3$\times$\textbf{conv}\\5-1\\ 512 \\$\mathrm{relu}$\\$T$-64} &
\tabincell{c}{\textbf{conv}\\25-1\\ 4096 \\$\mathrm{relu}$\\ $T$-64} &
\tabincell{c}{\textbf{conv}\\1-1\\ 4096 \\$\mathrm{relu}$\\ $T$-64} &
\tabincell{c}{\textbf{conv}\\1-1\\ $L$ \\$\mathrm{sigm}$\\ $T$-512} &
\tabincell{c}{\textbf{lconv-3D}\\3-50-1\\$L$\\ $\mathrm{lin}$\\$T$-512} &
\tabincell{c}{\textbf{conv-3D}\\3-9-1\\$L\times k$\\ $\mathrm{lin}$\\$T$-512} &
\tabincell{c}{\textbf{bmin} \\ 1-1 \\ $L$ \\ $\mathrm{idn}$\\$T$-512} &
\tabincell{c}{\textbf{sum} \\ 1-1 \\ $L$ \\ $\mathrm{soft}$\\$T$-512}\\
\hline
\end{tabular}
\vspace{2pt}
\begin{tablenotes}
As shown in (a) and (b) respectively. Each table contains five rows, `\textbf{layer}', `\textbf{fi.-st.}', `\textbf{\#ch.}',  `\textbf{act.}', and `\textbf{size}' represent the `\textbf{name of layer}', `\textbf{receptive field of filter}'$-$`\textbf{stride}', `\textbf{number of output feature maps}', `\textbf{activation function}', and `\textbf{size of output feature maps}', respectively. Furthermore, `\textbf{conv}',  `\textbf{lconv-3D}', `\textbf{conv-3D}', `\textbf{max}', `\textbf{bmin}', `\textbf{fc}', and `\textbf{sum}' represent the convolution, 3D convolution, 3D local convolution, max pooling, block min pooling, fully connection, and summation, respectively. Moreover, `relu', `idn', `soft', `sigm', and `lin' represent the activation functions, including rectified linear unit \cite{krizhevsky2012imagenet}, identity, softmax, sigmoid, and linear, respectively. $T$, $L$, and $k$ represent the length of frames, number of categories, and number of mixture filters.
\end{tablenotes}
\end{center}
\vspace{-10pt}
\end{table*}

%
This section describes the implementation of Eqn.~(\ref{eq:newq}) in a Deep Parsing Network (DPN).
DPN extends \vgg~to model the unary term and with additional layers to approximate one iteration of MF inference as the pairwise term.
%
%
The hyper-parameters of \vgg~and DPN are compared in Table \ref{tab:net}.

As listed in Table \ref{tab:net}, the first row represents the \emph{name} of layer
and `$x$-$y$' in the second row represents the \emph{size} of the receptive field and the \emph{stride} of convolution, respectively.
For instance, `3-1' in the convolutional layer implies that the receptive field of each filter is 3$\times$3 and it is applied on every single pixel of an input feature map, while `2-2' in the max-pooling layer indicates each feature map is pooled over every other pixel within a 2$\times$2 local region.
On the other hand, `3-50-1' implies a 3D convolution of size 50$\times$50$\times$3, where `3' suggests that this 3D filter is applied on three consecutive frames.
The last three rows show the number of the output feature maps, activation functions, and the size of output feature maps, respectively.
$T$ represents the number of frames in the underlying video sequence.

As summarized in Table \ref{tab:net}(a), \vgg~contains thirteen convolutional layers, five max-pooling layers, and three fully-connected layers.
These layers can be partitioned into twelve groups, each of which covers one or more homogeneous layers.
For example, the first group comprises two convolutional layers with 3$\times$3 receptive field and 64 output feature maps, each of which is 224$\times$224.

\subsection{Modeling Unary Terms}

To make full use of \vgg, which is pre-trained by ImageNet,
we adopt all its parameters to initialize the filters of the first ten groups of DPN. To simplify the discussions, we take PASCAL VOC 2012 (VOC12) \cite{everingham2010pascal} as an example. Note that DPN can be easily adapted to any other semantic image segmentation dataset by modifying its hyper-parameters.
VOC12 contains 21 categories and each image is rescaled to 512$\times$512 in training. Therefore, DPN needs to predict a total of 512$\times$512$\times$21 labels, \ie one label for each pixel.
To this end, we
extends \vgg~in two aspects.

\begin{figure}[t]
  \centering
  \includegraphics[width=0.45\textwidth]{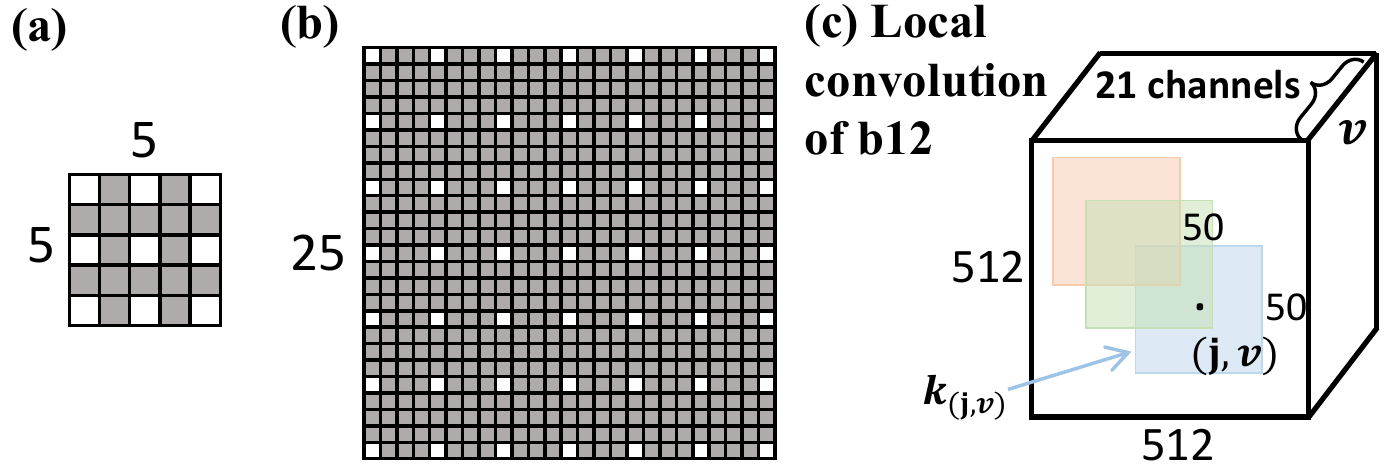}
  \caption{(a) and (b) show the padding of the filters. (c) illustrates local convolution of b12.}
  \label{fig:filter}
  \vspace{-8pt}
\end{figure}

In particular, let a$i$ and b$i$ denote the $i$-th group in Table \ref{tab:net}(a) and (b), respectively.
First, we \textbf{increase the resolution} of \vgg~by removing its max pooling layers at a8 and a10, since most of the information is lost after pooling, \eg a10 reduces the input size by 32 times, \ie from 224$\times$224 to 7$\times$7.
As a result, the smallest size of feature map in DPN is 64$\times$64, keeping much more information compared with \vgg.
Note that the filters of b8 are initialized as the filters of a9, but the 3$\times$3 receptive field is padded into 5$\times$5 as shown in Fig.~\ref{fig:filter}(a), where the cells in white are the original values of the a9's filter and the cells in gray are zeros.
This step is performed because a8 is not presented in DPN, therefore each filter in a9 should be convolved on every other pixel of a7.
To maintain the convolution with one stride, we pad the filters with zeros.
Furthermore, the feature maps in b11 are up-sampled to 512$\times$512 by bilinear interpolation.
Since DPN is trained with label maps of the entire images, the missing information in the preceding layers of b11 can be recovered by BP.
The supervision signals in the interpolated pixels will guide the feature learning for full-resolution images.

Second, two fully-connected layers at a11 are transformed into two convolutional layers at b9 and b10, respectively.
As shown in Table \ref{tab:net}(a), the first `fc' layer learns 7$\times$7$\times$512$\times$4096 parameters, which can be altered to 4096 filters in b9, each of which is 25$\times$25$\times$512.
%
Since a8 and a10 have been removed, the 7$\times$7 receptive field is padded into 25$\times$25 similar as above and shown in Fig.\ref{fig:filter} (b).
The second `fc' layer learns a 4096$\times$4096 weight matrix, corresponding to 4096 filters in b10. Each filter is 1$\times$1$\times$4096.

Overall, b11 generates the unary labeling results, producing twenty-one 512$\times$512 feature maps, each of which represents the probabilistic label map of each category.

\subsection{Modeling Smoothness Terms}
The last four layers of DPN, \ie from b12 to b15, are carefully designed to smooth the unary labeling results.

$\bullet$
\textbf{b12.} As listed in Table \ref{tab:net} (b), `lconv' in b12 indicates a \textbf{3D locally convolutional layer}.
A counterpart of it (\ie 2D locally convolutional layer) is widely used in face recognition \cite{sun2014deep, taigman2014deepface, liu2015deep} to capture different information from different facial positions.
Similarly, distinct spatial positions of b12 have different filters, and each filter is shared across 21 input channels, as shown in Fig.~\ref{fig:filter}(c).
It can be formulated as
\begin{equation}\label{eq:lconv}
\bo^{12}_{(\mathbf{j},v)}=\mathrm{lin}(\bk_{(\mathbf{j},v)}\ast\bo^{11}_{(\mathbf{j},v)}),
\end{equation}
where $\mathrm{lin}(x)=ax+b$ representing the linear activation function, `$\ast$' is the convolutional operator, and $\bk_{(\mathbf{j},v)}$ is a 50$\times$50$\times$3$\times$1 filter at position $\mathbf{j}$ of channel $v$.
The choice and effect of filter size will be discussed in the experiments (Sec.\ref{sec:ablation}).
We have $\bk_{(\mathbf{j},1)}=\bk_{(\mathbf{j},2)}=...=\bk_{(\mathbf{j},21)}$ shared across 21 channels.
$\bo^{11}_{(\mathbf{j},v)}$ indicates a local cube in b11, while $\bo^{12}_{(\mathbf{j},v)}$ is the corresponding output of b12.
%
%
Since b12 has a stride of one, the result of $\bk_{\mathbf{j}}\ast\bo^{11}_{(\mathbf{j},v)}$ is scalar.
In summary, b12 has 512$\times$512 different filters and produces 21 output feature maps.

Eqn.~(\ref{eq:lconv}) implements the \textbf{triple penalty} of Eqn.~(\ref{eq:newq}).
Recall that each output feature map of b11 indicates a probabilistic label map of a specific object appearing in the frame.
As a result, Eqn.~(\ref{eq:lconv}) suggests that the probability of object $v$ presented at position $\mathbf{j}$ is updated by weighted averaging over the probabilities at its nearby positions.
Thus, as shown in Fig.~\ref{fig:mrf}(c), $\bo^{11}_{(\mathbf{j},v)}$ corresponds to a cube of $Q^v$ centered at $\mathbf{j}$, which has values $p^v_{\mathbf{z}}$, $\forall \mathbf{z}\in\mathcal{N}_{\mathbf{j}}^{50\times50\times3}$.
Similarly, $\bk_{(\mathbf{j},v)}$ is initialized by $\mathrm{d}(\mathbf{j},\mathbf{z})p^v_{\mathbf{j}}$, implying each filter captures dissimilarities between positions.
These filters remain fixed during BP,
other than learned as in conventional CNN\footnote{Each filter in b12 actually represents a distance metric between pixels in a specific region.
In VOC12, the patterns of all the training images in a specific region are heterogeneous, because of various object shapes.
Therefore, we initialize each filter with Euclidean distance.
Nevertheless, Eqn.~(\ref{eq:lconv}) is a more general form than the triple penalty in Eqn.~(\ref{eq:newq}), \ie filters in Eqn.~(\ref{eq:lconv}) can be automatically learned from data,
if the patterns in a specific region are homogeneous, such as face or human images, which have more regular shapes than images in VOC12.}.

\begin{figure}[t]
  \centering
  \includegraphics[width=0.45\textwidth]{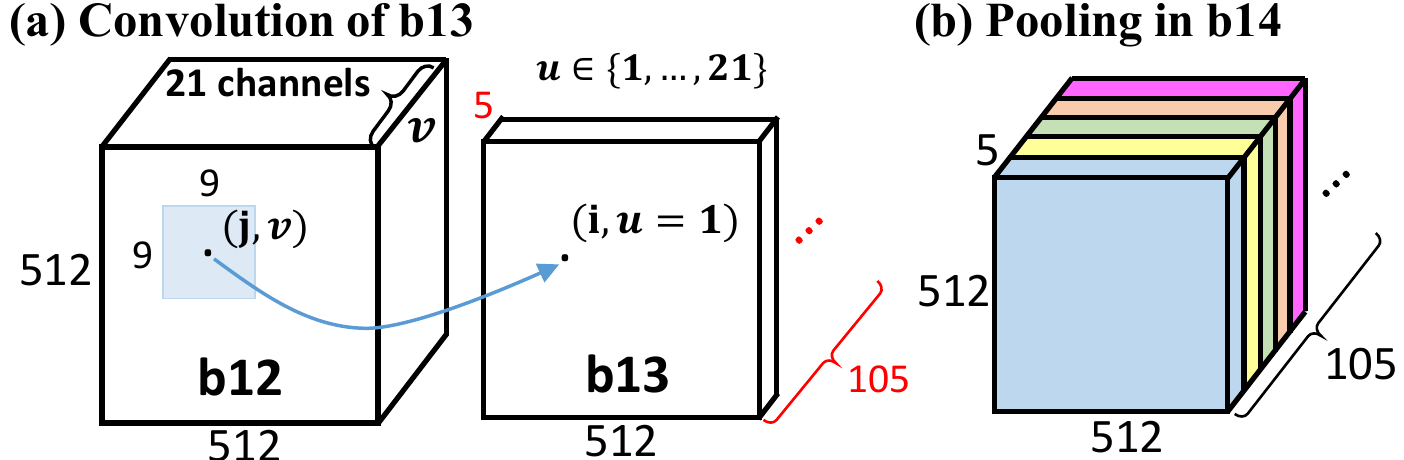}
  \caption{(a) and (b) illustrates the convolutions of b13 and the poolings in b14.}
  \label{fig:b1314}
  \vspace{-8pt}
\end{figure}

$\bullet$ \textbf{b13.}
As shown in Table \ref{tab:net}(b) and Fig.~\ref{fig:b1314}(a), b13 is a \textbf{3D global convolutional layer} that generates 105 feature maps by using 105 filters of size 9$\times$9$\times$3$\times$21.
For example, the value of $(\mathbf{i},u=1)$ is attained by applying a 9$\times$9$\times$3$\times$21 filter at positions $\{(\mathbf{j},v=1,...,21)\}$.
In other words, b13 learns a filter for each category to penalize the probabilistic label maps of b12, corresponding to the \textbf{local label contexts} in Eqn.(\ref{eq:newq}) by assuming $K=5$ and $n=9$, as shown in Fig.\ref{fig:mrf} (d).
%

%
%

$\bullet$ \textbf{b14.} As illustrated in Table \ref{tab:net} and Fig.~\ref{fig:b1314}(b), b14 is a block min pooling layer that pools over every 1$\times$1 region with one stride across every 5 input channels, leading to 21 output channels, \ie 105$\div$5$=$21.
Layer b14 activates the contextual pattern with the smallest penalty.

$\bullet$ \textbf{b15.} This layer combines both the unary and smoothness terms by summing the outputs of b11 and b14 in an element-wise manner similar to Eqn.~(\ref{eq:newq}),
\begin{equation}
\bo^{15}_{(\mathbf{i},u)}=\frac{\exp\big\{\ln(\bo^{11}_{(\mathbf{i},u)})-\bo^{14}_{(\mathbf{i},u)}\big\}}
{\sum_{u=1}^{21}\exp\big\{
\ln(\bo^{11}_{(\mathbf{i},u)})-\bo^{14}_{(\mathbf{i},u)}\big\}},
\end{equation}
where probability of assigning label $u$ to voxel $\mathbf{i}$ is normalized over all the labels.

\vspace{5pt}
\vspace{4pt}\hspace{-15pt}\textbf{Relation to Previous Deep Models.~}
Many existing deep models such as \cite{zheng2015conditional, chen2014semantic, schwing2015fully} employed Eqn.~(\ref{eq:old_p}) as the pairwise terms,
which are the special cases of Eqn.~(\ref{eq:newq}).
To see this, let $K=1$, $j=i$ and omit $t$ (\ie $t_{z}=t_{j}=t_{i}$), the right hand side of Eqn.~(\ref{eq:newq}) reduces to
%
%
%
\begin{eqnarray}\label{eq:K1s1}
&&\exp\{-\Phi_i^u-\sum_{v\in L}\lambda_1\mu_1(i,u,i,v)\sum_{z\in\mathcal{N}_i} \mathrm{d}(i,z)p_i^vp_z^v\}\nonumber\\
&=&\exp\{-\Phi_i^u-\sum_{v\in L}\mu(u,v)\sum_{z\in\mathcal{N}_i,z\neq i} \mathrm{d}(i,z)p_z^v\},
\end{eqnarray}
where $\mu(u,v)$ and $\mathrm{d}(i,z)$ represent the global label co-occurrence and pairwise pixel similarity of Eqn.~(\ref{eq:old_p}), respectively.
This is because $\lambda_1$ is a constant, $\mathrm{d}(i,i)=0$, and $\mu(i,u,i,v)=\mu(u,v)$.
%
Eqn.~(\ref{eq:K1s1}) is the corresponding MF update equation of (\ref{eq:old_p}).

\subsection{Learning Algorithms}\label{sec:learn}

We describe the training strategy of DPN and also its time complexity and efficient implementation.

\vspace{4pt}\hspace{-15pt}\textbf{Learning.~} The first ten groups of DPN are initialized by \vgg\footnote{We use the released \vgg~model, which is publicly available at \url{http://www.robots.ox.ac.uk/~vgg/research/very_deep/}}, while the last four groups can be initialized randomly.
DPN is then fine-tuned in an incremental manner with four stages.
%
During fine-tuning, all these stages solve the pixelwise softmax loss \cite{long2014fully}, but updating different sets of parameters.

First, we add a loss function to b11 and fine-tune the weights from b1 to b11 without the last four groups, in order to learn the unary terms.
Second, to learn the triple relations, we stack b12 on top of b11 and update its parameters (\ie $\omega_1,\omega_2$ in the distance measure), but the weights of the preceding groups (\ie b1$\sim$b11) are fixed.
Third, b13 and b14 are stacked onto b12 and similarly, their weights are updated with all the preceding parameters fixed, so as to learn the local label contexts.
Finally, all the parameters are jointly fine-tuned.
%

\vspace{4pt}\hspace{-15pt}\textbf{Complexity.~} DPN transforms Eqn.~(\ref{eq:newq}) into convolutions and poolings in the groups from b12 to b15, such that filtering at each pixel can be performed in a parallel manner.
Assume we have $f$ input and $f'$ output feature maps, $N\times N$ pixels, filters with $s\times s$ receptive field, and a mini-batch with $M$ samples.
b12 takes a total $f\cdot N^2\cdot s^2\cdot M$ operations, b13 takes $f\cdot f'\cdot N^2\cdot s^2\cdot M$ operations, while both b14 and b15 require $f\cdot N^2\cdot M$ operations.
For example, when $M$$=$10 as in our experiment, we have 21$\times$512$^2$$\times$50$^2$$\times$10$=$1.3$\times$10$^{11}$ operations in b12, which has the highest complexity in DPN.
If we parallelize these operations using matrix multiplication on GPU as \cite{chetlur2014cudnn} did, the operation in b12 can be computed within 30ms.
The total runtime of the last four layers of DPN is 75ms.
Note that convolutions in DPN can be further speeded up by low-rank decompositions \cite{jaderberg2014speeding} of the filters and model compressions \cite{hinton2014distilling}.

In contrast, existing works \cite{chen2014semantic, zheng2015conditional} employ fast Gaussian filtering \cite{adams2010fast} to accelerate the direct calculation of Eqn.~(\ref{eq:newq}).
%
{For a mini-batch of ten 512$\times$512 images, a recently optimized implementation \cite{koltun2011efficient} needs 3.5$\times$10$^{11}$ operations and takes 114ms on GPU to compute one iteration of (\ref{eq:newq}).}
%
In DPN, simultaneous message passing is enabled over every training image, in the sense that parallelization is obtained in image level.
Therefore, DPN makes (\ref{eq:newq}) easier to be parallelized and speeded up.

\vspace{4pt}\hspace{-15pt}\textbf{Efficient Implementation.~}
%
As mentioned in Eqn.~(\ref{eq:old_p}), the local filters in b12 are computed by the distances between RGB values of the pixels.
XY coordinates are omitted here because they could be pre-computed.
To accelerate the computation of local convolution, the lookup table-based filtering approach is employed.
Specifically, we first construct a lookup table storing distances between any two pixel intensities (ranging from 0 to 255),
which results in a $256 \times 256$ matrix.
Then when we perform local convolution, the kernels' coefficients can be obtained efficiently by just looking up the table.



\section{Experiments}

In this section, we demonstrate the effectiveness of DPN and benchmark it against other state-of-the-art semantic segmentation methods.
Below, we give an overview of the dataset and evaluation metrics used in these experiments.
Representative methods are also introduced.
In the following experiments, we denote $2$-D DPN as DPN and $3$-D DPN as spatial-temporal DPN.

\vspace{4pt}\hspace{-15pt}\textbf{Dataset.~} We compare DPN with the state-of-the-art methods on PASCAL VOC 2012 (VOC12) \cite{everingham2010pascal} , Cityscapes \cite{Cordts2016Cityscapes} and CamVid \cite{BrostowSFC:ECCV08} datasets. 
VOC12 is a well-known benchmark for generic image segmentation and Cityscapes dataset focuses on parsing urban street scenes.
We choose those two benchmarks to evaluate the original DPN. 
%
%
On the other hand, CamVid dataset is composed of several video sequences, which is suitable for the evaluation of spatial-temporal DPN.
We summarize the information of all datasets we used in Table \ref{tab:dataset}.

\begin{table}
	\small
	\caption{Summary of datasets.}
	\label{tab:dataset}
	\centering
	\begin{tabular}{c|c|c|c|c}
		\hline
		name&training&validation&testing&video data\\
		\hline\hline
		VOC12&10582&1449&1456&no\\
		Cityscapes&2975&500&1525&no\\
		CamVid&367&-&233&yes\\
		\hline
	\end{tabular}
\end{table}

\vspace{4pt}\hspace{-15pt}\textbf{Evaluation Metrics.~}
All existing works employed mean pixelwise intersection-over-union (denoted as mIoU) \cite{long2014fully} to evaluate their performance.
To fully examine the effectiveness of DPN, we introduce another three metrics, including tagging accuracy (TA), localization accuracy (LA), and boundary accuracy (BA).
(1) TA compares the predicted image-level tags with the ground truth tags, calculating the accuracy of multi-class image classification.
%
%
(2) LA evaluates the IoU between the predicted object bounding boxes\footnote{They are the bounding boxes of the predicted segmentation regions.} and the ground truth bounding boxes (denoted as bIoU), measuring the precision of object localization.
(3) For those objects that have been correctly localized, we compare the predicted object boundary with the ground truth boundary,
measuring the precision of semantic boundary similar to \cite{hariharan2011semantic}.

\vspace{4pt}\hspace{-15pt}\textbf{Comparisons.~}
%
%
DPN is compared with the state-of-the-art segmentation methods, including FCN \cite{long2014fully}, Zoom-out \cite{mostajabi2014feedforward}, DeepLab \cite{chen2014semantic}, WSSL \cite{papandreou2015weakly}, BoxSup \cite{dai2015boxsup}, Piecewise \cite{lin2015efficient}, RNN \cite{zheng2015conditional}, SuperParsing \cite{tighe2010superparsing}, Dilation10 \cite{yu2015multi}, ALE \cite{russell2009associative}, Multiclass \cite{liu2015multiclass}, and SegNet \cite{badrinarayanan2015segnet}.
All these methods are based on CNNs or MRFs. They can be grouped according to different aspects:
(1) \textbf{joint-train}: Piecewise and RNN; (2) \textbf{w/o joint-train}: DeepLab, WSSL, FCN, and BoxSup; (3) \textbf{pre-train on COCO}: RNN, WSSL, and BoxSup.
The first and the second groups are the methods with and without joint training CNNs and MRFs, respectively.
%
%
Methods in the last group also employed MS-COCO \cite{lin2014microsoft} to pre-train deep models.
To conduct a comprehensive comparison, the performance of DPN are reported on both settings, \ie, with and without pre-training on COCO.

In the following, Sec.~\ref{sec:ablation} investigates the effectiveness of different components of DPN on the VOC12.
Sec.~\ref{sec:st-ablation} evaluates the spatial-temporal DPN on the CamVid.
Sec.~\ref{sec:analysis} provides detailed analysis of the DPN system and its performance. 
Sec.~\ref{sec:overall} compares DPN with the state-of-the-art methods on the benchmarks.

\subsection{Effectiveness of DPN}
\label{sec:ablation}

All the models evaluated in this section are trained on the \emph{training set} and tested on the \emph{validation set} of VOC12. 
%

\vspace{4pt}\hspace{-15pt}\textbf{Triple Penalty.~} The receptive field of b12 indicates the range of triple relations for each pixel.
We examine different settings of the receptive fields, including `10$\times$10', `50$\times$50', and `100$\times$100', as shown in Table \ref{tab:ablation}(a), where
`50$\times$50' achieves the best mIoU, which is sightly better than `100$\times$100'.
For a 512$\times$512 image, this result implies that 50$\times$50 neighborhood is sufficient to capture relations between pixels, while smaller or larger regions tend to under-fit or over-fit the training data.
Moreover, all models of triple relations outperform the `baseline' method that models dense pairwise relations, \ie \vgg+denseCRF \cite{koltun2011efficient}.

\begin{table}[t]
\small
\caption{Ablation study of hyper-parameters.}
\label{tab:ablation}
\begin{subtable}{\linewidth}
\centering
\begin{tabular}{c|c|c|c|c}
\hline
Receptive Field & baseline & 10$\times$10 & 50$\times$50 & 100$\times$100 \\
\hline\hline
mIoU (\%) & 63.4 & 63.8 & \textbf{64.7} & 64.3 \\
\hline
\end{tabular}\vspace{5pt}
\subcaption{\footnotesize Comparisons between different receptive fields of b12.}
\end{subtable}

\begin{subtable}{\linewidth}
\centering
\begin{tabular}{c|c|c|c|c}
\hline
Receptive Field & 1$\times$1 & 5$\times$5 & 9$\times$9 & 9$\times$9 mixtures \\
\hline\hline
mIoU (\%) & 64.8 & 66.0 & 66.3 & \textbf{66.5} \\
\hline
\end{tabular}\vspace{5pt}
\subcaption{\footnotesize Comparisons between different receptive fields of b13.}
\end{subtable}

\begin{subtable}{\linewidth}
\centering
\begin{tabular}{c|c|c|c}
\hline
Pairwise Terms & DSN~\cite{schwing2015fully} & DeepLab~\cite{chen2014semantic} & DPN \\
\hline\hline
improvement (\%) & 2.6 & 3.3 & \textbf{5.4} \\
\hline
\end{tabular}\vspace{5pt}
\subcaption{\footnotesize Comparing pairwise terms of different methods.}
\end{subtable}
\end{table}

\begin{figure}[t]
  \centering
  \includegraphics[width=0.48\textwidth]{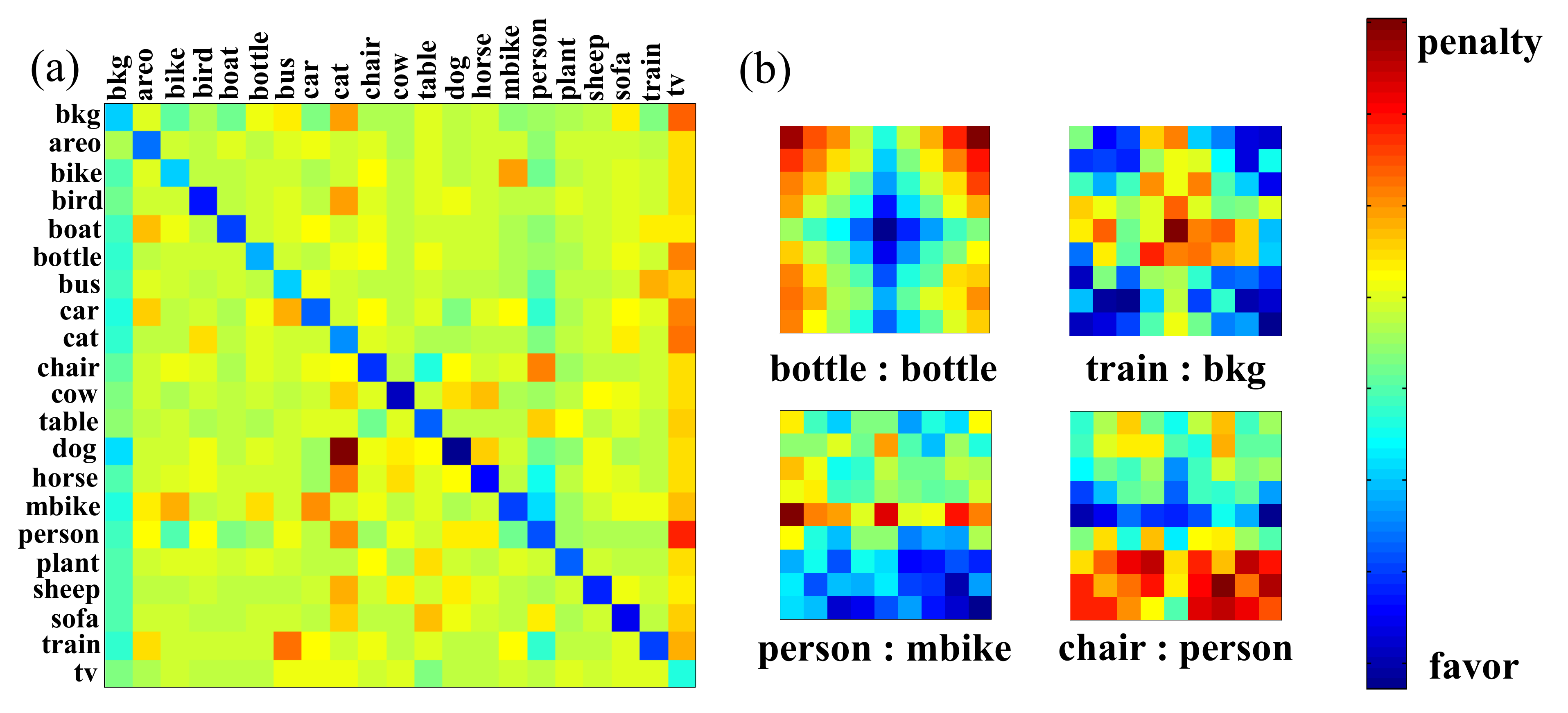}
  \caption{Visualization of (a) learned label compatibility (b) learned contextual information. \textbf{(Best viewed in color)}}
  \label{fig:label}
\end{figure}

\vspace{4pt}\hspace{-15pt}\textbf{Label Contexts.~} Receptive field of b13 indicates the range of local label context.
To evaluate its effectiveness, we fix the receptive field of b12 as 50$\times$50.
As summarized in Table \ref{tab:ablation}(b), `9$\times$9 mixtures' improves preceding settings by 1.7, 0.5, and 0.2 percent respectively.
We observe large gaps exist between `1$\times$1' and `5$\times$5'.
Note that the 1$\times$1 receptive field of b13 corresponds to learning a global label co-occurrence without considering local spatial contexts.
%
%
Table \ref{tab:ablation}(c) shows that the pairwise terms of DPN are more effective than DSN and DeepLab\footnote{The other deep models such as RNN and Piecewise did not report the exact improvements after combining unary and pairwise terms.}.

More importantly, mIoU of all the categories can be improved through increasing the size of receptive field and learning a mixture.
Specifically, for each category, the improvements of the last three settings in Table \ref{tab:ablation}(b) over the first one are 1.2$\pm$0.2, 1.5$\pm$0.2, and 1.7$\pm$0.3, respectively.

We also visualize the learned label compatibilities and contexts in Fig.~\ref{fig:label}(a) and (b), respectively.
Fig.~\ref{fig:label}(a) is obtained by summing each filter in b13 over a 9$\times$9 region, indicating how likely a column object would present when a row object is presented. Blue color represents high favorability.
It is worth pointing out that Fig.~\ref{fig:label}(a) is non-symmetry.
For example, when a `horse' is presented, a `person' is more likely to present than the other objects. On the other hand, when a `person' presents in an image, it is more likely to find a `bike' (than a `horse'). The `bkg' (background) is compatible with all the objects.
Fig.~\ref{fig:label}(b) visualizes some contextual patterns, where `A:B' indicates that when `A' is presented, where `B' is more likely to present.
For example, `bkg' is around `train', `motor bike' is below `person', and `person' is sitting on `chair'.

\begin{table}
\small
\caption{The effectiveness of spatial-temporal DPN.}
\label{tab:st-ablation}
	\centering
	\begin{tabular}{l|c|c}
	  \hline
	  &2D&3D\\
	  \hline\hline
	  Unary Term (mIoU)&59.67&59.67\\
	  ~~+ Triple Penalty&59.93&\textbf{60.16}\\
	  ~~+ Label Contexts&60.06&\textbf{60.25}\\
	  \hline
	\end{tabular}
\end{table}

\begin{figure}[t]
	\centering
	\includegraphics[width=0.48\textwidth]{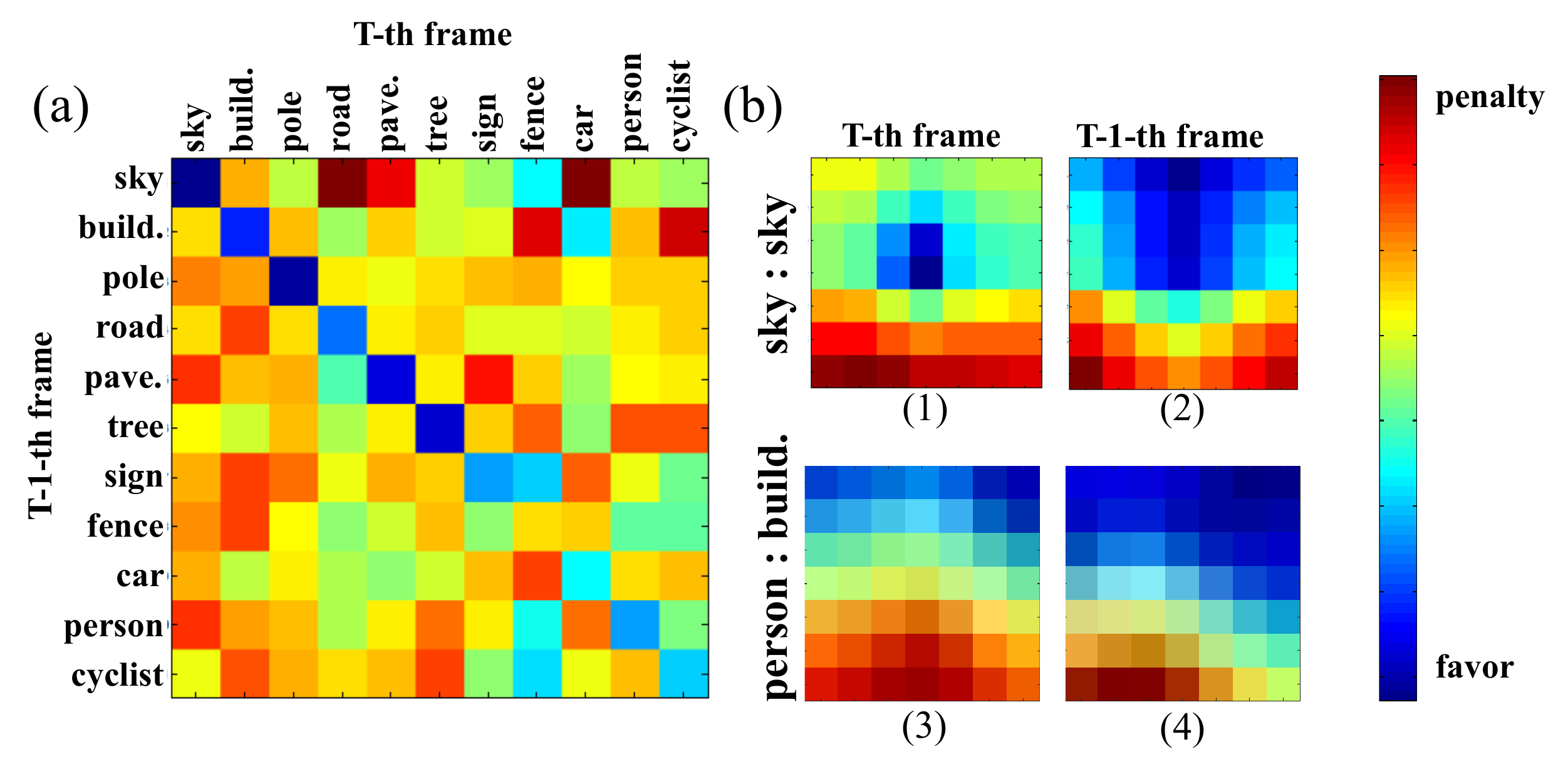}
	\caption{Visualization of (a) 3D learned label compatibility (b) learned spatial-temporal contextual information. \textbf{(Best viewed in color)}}
	\label{fig:label3D}
\end{figure}

\begin{figure}[t]
	\centering
	\includegraphics[width=0.45\textwidth]{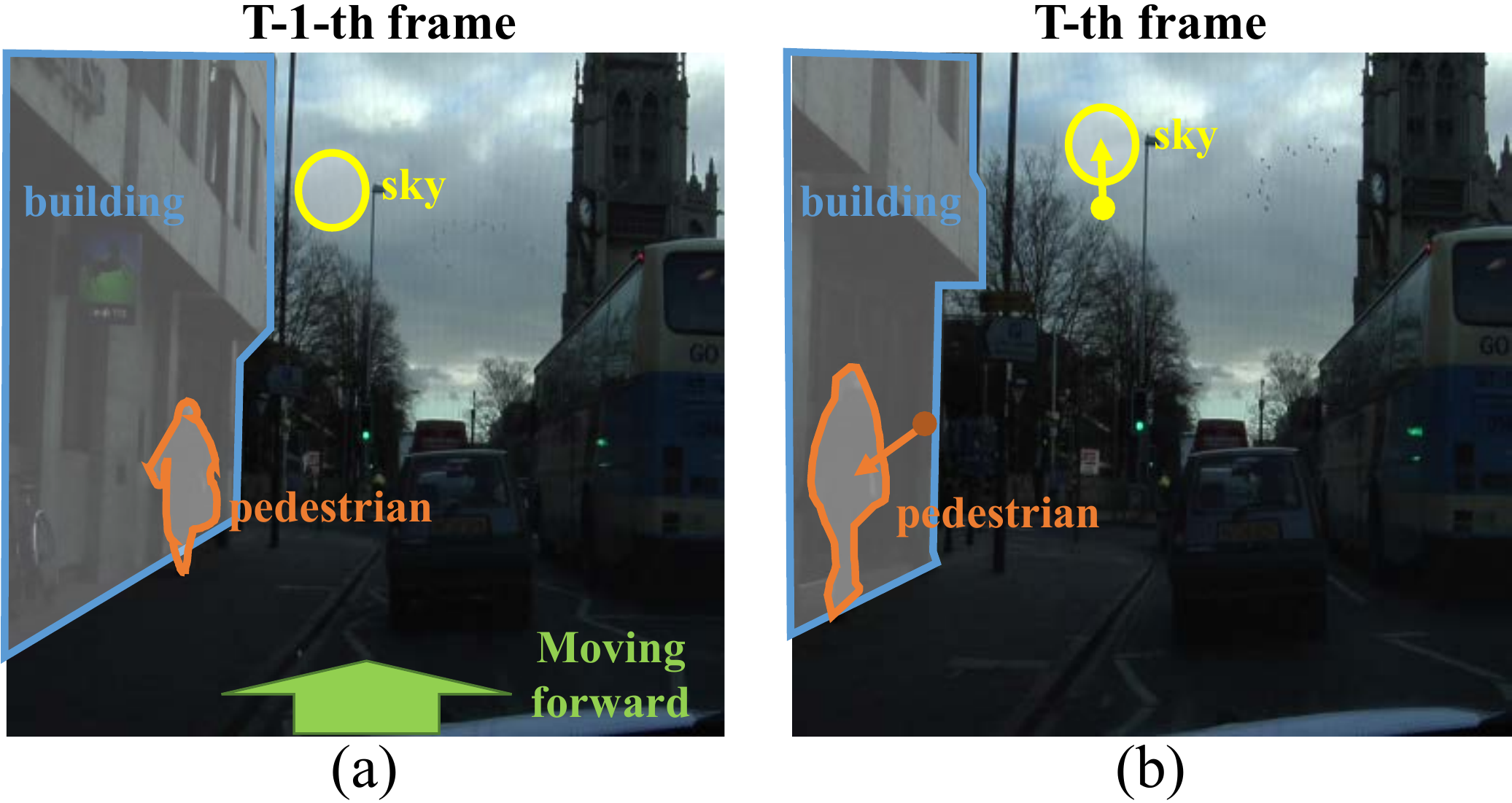}
	\caption{(a) and (b) are neighboring video frames from CamVid dataset. When the vehicle moved forward, the `sky' (blue) shift upward and the `pedestrian' (orange) shift downward. \textbf{(Best viewed in color)}}
	\label{fig:movement}
\end{figure}

\subsection{Effectiveness of Spatial-temporal DPN}
\label{sec:st-ablation}
In this section, we quantitatively investigate the effectiveness of the pairwise term in spatial-temporal DPN.
CamVid {\em train} are used for training and performance are reported on CamVid {\em test}.
Images in CamVid are captured from a moving vehicle, thus exhibiting certain temporal regularization.

\vspace{4pt}\hspace{-15pt}\textbf{3D Convolution.~}
In spatial-temporal DPN, the b12 is a 3D local convolutional layer and b13 is a 3D global convolutional layer. In Table \ref{tab:st-ablation}, we evaluate the performance gain of each stage on DPN. According to Table \ref{tab:ablation}(a), we set the receptive field of b12 as 50$\times$50 and b13 as 9$\times$9 in 2D setting. In 3D pairwise term, the receptive field of b12 is 50$\times$50$\times$3 and b13 is 7$\times$7$\times$3. We observe that applying 2D and 3D pairwise terms on unary term both improves the performance. Since 3D pairwise terms capture the information between successive frames, it performs slightly better than 2D pairwise terms.

\vspace{4pt}\hspace{-15pt}\textbf{Temporal Regularization.~}
In Fig.~\ref{fig:label3D}, we visualize the 3D learned label compatibilities and contexts across frames.

Fig.~\ref{fig:label3D}(a) depicts the possibility of column object would present in the $T$-${th}$ frame when a row object is presented in the $T-1$-${th}$ frame. Blue color indicates high favorability. For instance, when `pavement' is presented in the $T-1$-${th}$ frame, `road' is more likely to present around in the $T$-${th}$ frame. However, if `sky' is presented in the $T-1$-${th}$ frame, the pairwise term will penalize the probability which `road' is presented in the $T$-${th}$ frame.


Fig.~\ref{fig:label3D}(b) visualizes the spatial$-$temporal contextual patterns. In each row,  `A:B' indicates how `A' influences `B'.The figures suggest that when `A' is presented in the $T$-${th}$ or the $T-1$-${th}$ frame, where `B' is more likely to present in the $T$-${th}$ frame.
In the first row, (1) represents that when `sky' is presented in the $T$-${th}$ frame, its neighborhood is more likely to be `sky'. (2) means when `sky' is presented in the $T-1$-${th}$ frame, it is more likely presented above this position in the $T$-${th}$ frame. 
As shown in Fig.~\ref{fig:movement}, (a) and (b) are neighboring video frames, when the vehicle moves forward from (a) to (b), `sky' region (yellow) shifts upward in the scene. So compared with the favor region in  Fig.~\ref{fig:label3D}(b)-(1), the favor region also shifts upward in (2).
In the second row of Fig.~\ref{fig:label3D}(b), (3) and (4) represent that when `pedestrian' is presented in the  $T$-${th}$ or the $T-1$-${th}$ frame, `building' is more likely presented above in the $T$-${th}$ frame. We observe that the shifting from (3) to (4) is similar with the shifting of `pedestrian'(orange) in Fig.~\ref{fig:movement} from (a) to (b).

%

\begin{figure}[t]
  \centering
  \includegraphics[width=0.48\textwidth]{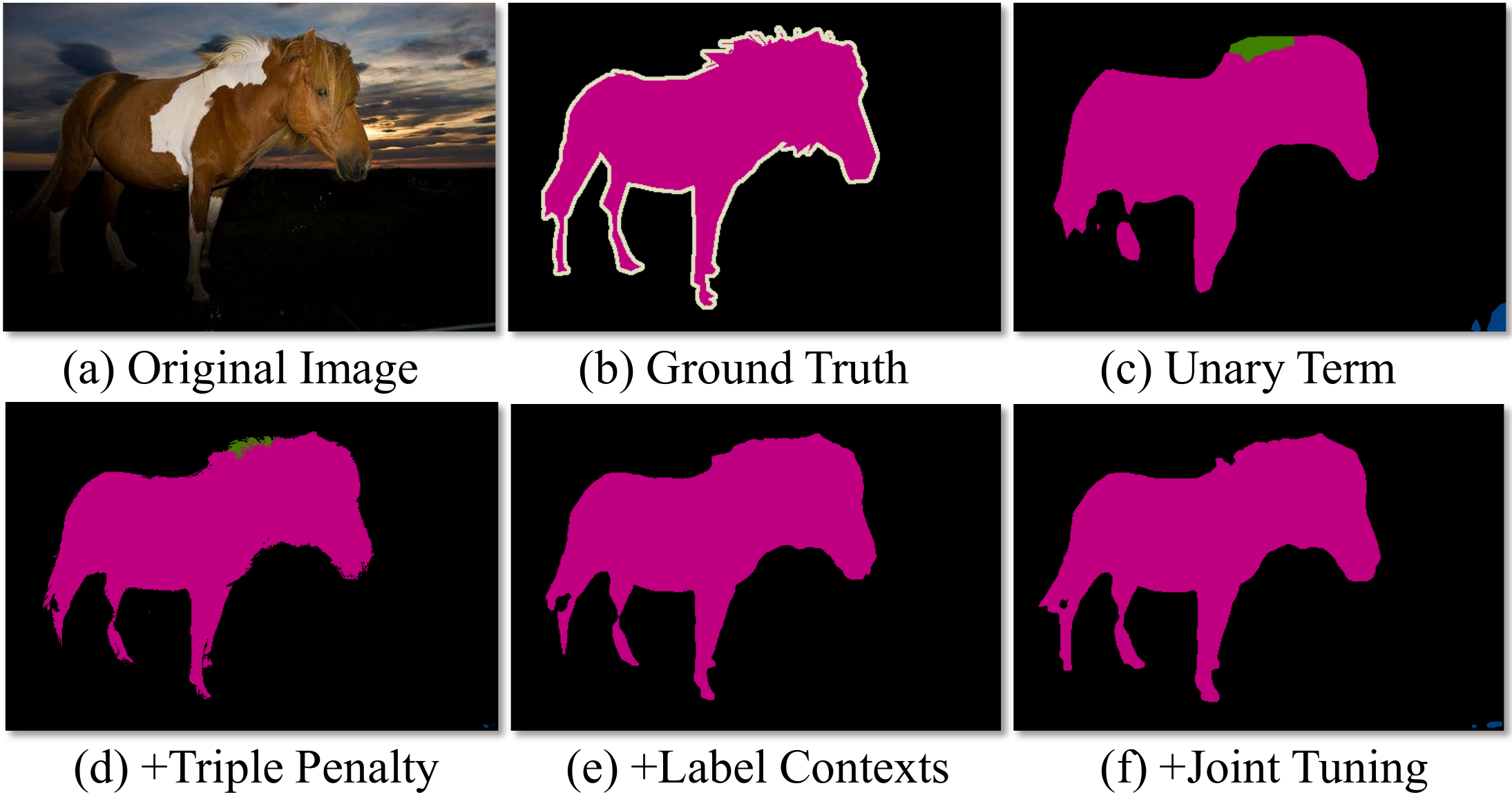}
  \caption{Step-by-step visualization of DPN. \textbf{(Best viewed in color)}}
  \label{fig:vis_pipeline}
\end{figure}

\subsection{Further Analysis}
\label{sec:analysis}

All the analysis in this section are conducted on the \emph{validation set} of VOC12.

\vspace{4pt}\hspace{-15pt}\textbf{Incremental Learning.~} As discussed in Sec.~\ref{sec:learn}, DPN is trained in an incremental manner.
The right hand side of Table \ref{tab:perclass}(a) demonstrates that each stage leads to performance gain compared to its previous stage.
For instance, `triple penalty' improves `unary term' by 2.3 percent, while `label contexts' improves `triple penalty' by 1.8 percent.
More importantly, joint fine-tuning all the components (\ie unary terms and pairwise terms) in DPN achieves another gain of 1.3 percent.
A step-by-step visualization is provided in Fig.~\ref{fig:vis_pipeline}.

\begin{figure}[t]
  \centering
  \includegraphics[width=0.48\textwidth]{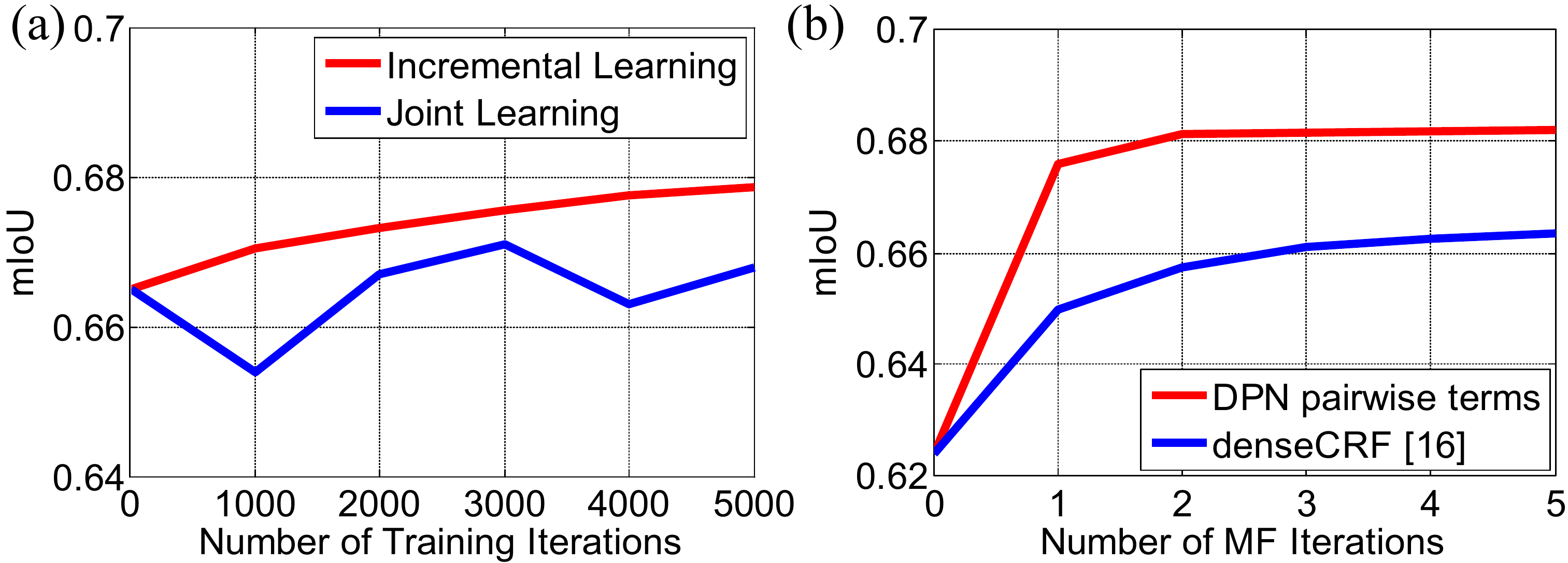}
  \caption{Ablation study of (a) training strategy (b) required MF iterations. \textbf{(Best viewed in color)}}
  \label{fig:strategy}
\end{figure}

We also compare `incremental learning' with `joint learning', which fine-tunes all the components of DPN at the same time.
The training curves of them are plotted in Fig.\ref{fig:strategy} (a), showing that the former leads to higher and more stable accuracies with respect to different iterations, while the latter may get stuck at local minima.
This difference is easy to understand -- incremental learning only introduces new parameters until all existing parameters have been fine-tuned.

\vspace{4pt}\hspace{-15pt}\textbf{One-iteration MF.~} DPN approximates one iteration of MF.
Fig.~\ref{fig:strategy}(b) illustrates that DPN reaches a good accuracy with one MF iteration.
%
%
A CRF \cite{koltun2011efficient} with dense pairwise edges needs more than 5 iterations to converge. It also has a large gap compared to DPN.
%
Note that the existing deep models such as \cite{chen2014semantic, zheng2015conditional, schwing2015fully} required 5$\sim$10 iterations to converge as well.

\begin{figure}[t]
  \centering
  \includegraphics[width=0.48\textwidth]{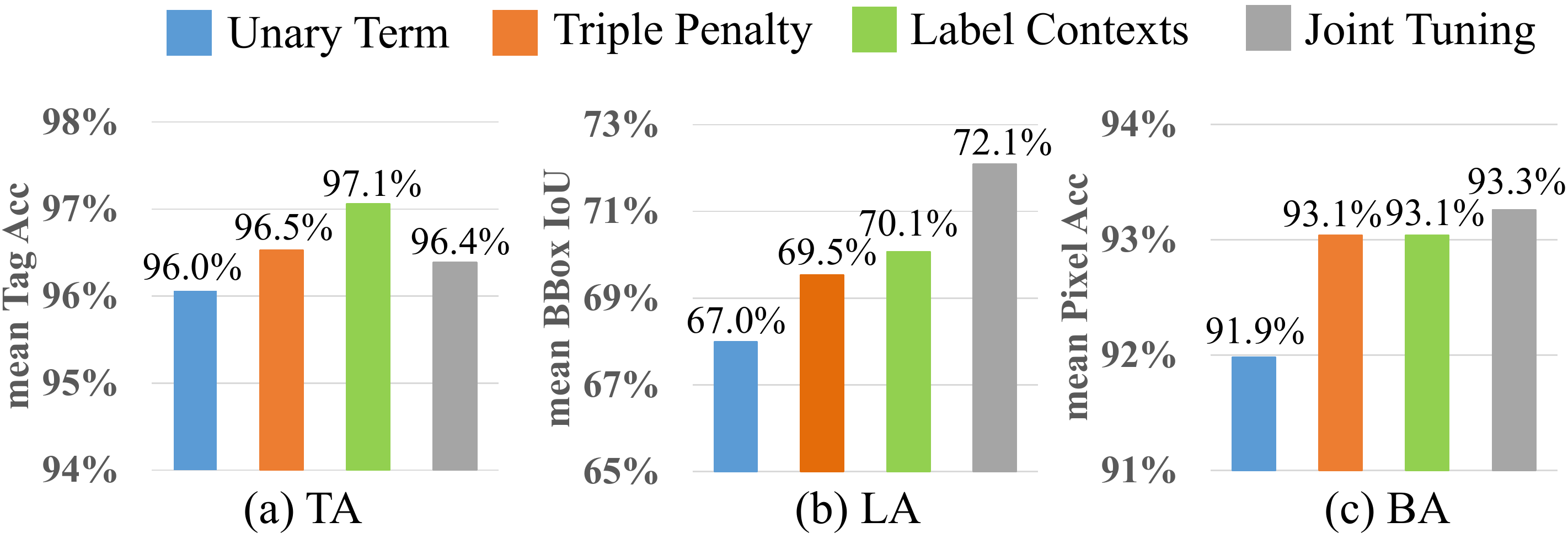}
  \caption{Stage-wise analysis of (a) mean tagging accuracy (b) mean localization accuracy (c) mean boundary accuracy.}
  \label{fig:error}
\end{figure}

\begin{table*}
\scriptsize
\caption{Per-class results on VOC12.}
\label{tab:perclass}
\begin{subtable}{\linewidth}
\centering
\begin{tabular}{l|p{9pt}p{9pt}p{9pt}p{9pt}p{9pt}p{9pt}p{9pt}p{9pt}p{9pt}p{9pt}p{9pt}p{9pt}p{9pt}p{9pt}p{9pt}p{9pt}p{9pt}p{9pt}p{9pt}p{9pt}|p{12pt}}
\hline
 & areo & bike & bird & boat & bottle & bus & car & cat & chair & cow & table & dog & horse & mbike & person & plant & sheep & sofa & train & tv & \textbf{Avg.} \\
\hline\hline
Unary Term (mIoU) & 77.5 & 34.1 & 76.2 & 58.3 & 63.3 & 78.1 & 72.5 & 76.5 & 26.6 & 59.9 & 40.8 & 70.0 & 62.9 & 69.3 & 76.3 & 39.2 & 70.4 & 37.6 & 72.5 & 57.3 & 62.4 \\
~~+ Triple Penalty & 82.3 & 35.9 & 80.6 & 60.1 & 64.8 & 79.5 & 74.1 & 80.9 & 27.9 & 63.5 & 40.4 & 73.8 & 66.7 & 70.8 & 79.0 & 42.0 & 74.1 & 39.1 & 73.2 & 58.5 & 64.7 \\
~~+ Label Contexts & 83.2 & 35.6 & \textbf{82.6} & 61.6 & 65.5 & 80.5 & 74.3 & \textbf{82.6} & 29.9 & \textbf{67.9} & 47.5 & 75.2 & \textbf{70.3} & 71.4 & 79.6 & 42.7 & \textbf{77.8} & 40.6 & 75.3 & 59.1 & 66.5 \\
~~+ Joint Tuning & \textbf{84.8} & \textbf{37.5} & 80.7 & \textbf{66.3} & \textbf{67.5} & \textbf{84.2} & \textbf{76.4} & 81.5 & \textbf{33.8} & 65.8 & \textbf{50.4} & \textbf{76.8} & 67.1 & \textbf{74.9} & \textbf{81.1} & \textbf{48.3} & 75.9 & \textbf{41.8} & \textbf{76.6} & \textbf{60.4} & \textbf{67.8} \\
\hline\hline
TA (tagging Acc.) & 98.8 & 97.9 & 98.4 & 97.7 & 96.1 & 98.6 & 95.2 & 96.8 & 90.1 & 97.5 & 95.7 & 96.7 & 96.3 & 98.1 & 93.3 & 96.1 & 98.7 & 92.2 & 97.4 & 96.3 & 96.4 \\
LA (bIoU)  & 81.7 & 76.3 & 75.5 & 70.3 & 54.4 & 86.4 & 70.6 & 85.6 & 51.8 & 79.6 & 57.1 & 83.3 & 79.2 & 80.0 & 74.1 & 53.1 & 79.1 & 68.4 & 76.3 & 58.8 & 72.1\\
BA (boundary Acc.) & 95.9 & 83.9 & 96.9 & 92.6 & 93.8 & 94.0 & 95.7 & 95.6 & 89.5 & 93.3 & 91.4 & 95.2 & 94.2 & 92.7 & 94.5 & 90.4 & 94.8 & 90.5 & 93.7 & 96.6 & 93.3 \\
\hline
\end{tabular}
\subcaption{\footnotesize Per-class results on VOC12 {\em val}.}
\end{subtable}

\begin{subtable}{\linewidth}
\centering
\begin{tabular}{l|p{9pt}p{9pt}p{9pt}p{9pt}p{9pt}p{9pt}p{9pt}p{9pt}p{9pt}p{9pt}p{9pt}p{9pt}p{9pt}p{9pt}p{9pt}p{9pt}p{9pt}p{9pt}p{9pt}p{9pt}|p{12pt}}
\hline
 & areo & bike & bird & boat & bottle & bus & car & cat & chair & cow & table & dog & horse & mbike & person & plant & sheep & sofa & train & tv & mIoU \\
\hline\hline
FCN \cite{long2014fully} & 76.8 & 34.2 & 68.9 & 49.4 & 60.3 & 75.3 & 74.7 & 77.6 & 21.4 & 62.5 & 46.8 & 71.8 & 63.9 & 76.5 & 73.9 & 45.2 & 72.4 & 37.4 & 70.9 & 55.1 & 62.2 \\
Zoom-out \cite{mostajabi2014feedforward} & 85.6 & 37.3 & 83.2 & 62.5 & 66.0 & 85.1 & 80.7 & 84.9 & 27.2 & 73.2 & 57.5 & 78.1 & 79.2 & 81.1 & 77.1 & 53.6 & 74.0 & 49.2 & 71.7 & 63.3 & 69.6 \\
DeepLab \cite{chen2014semantic} & 84.4 & 54.5 & 81.5 & 63.6 & 65.9 & 85.1 & 79.1 & 83.4 & 30.7 & 74.1 & 59.8 & 79.0 & 76.1 & 83.2 & 80.8 & 59.7 & 82.2 & 50.4 & 73.1 & 63.7 & 71.6 \\
RNN \cite{zheng2015conditional} & 87.5 & 39.0 & 79.7 & 64.2 & 68.3 & 87.6 & 80.8 & 84.4 & 30.4 & 78.2 & 60.4 & 80.5 & 77.8 & 83.1 & 80.6 & 59.5 & 82.8 & 47.8 & 78.3 & 67.1 & 72.0 \\
Piecewise \cite{lin2015efficient} & 90.6 & 37.6 & 80.0 & 67.8 & 74.4 & 92.0 & \textbf{85.2} & 86.2 & 39.1 & 81.2 & 58.9 & 83.8 & 83.9 & 84.3 & 84.8 & 62.1 & 83.2 & 58.2 & 80.8 & 72.3 & 75.3 \\
\hline
WSSL$^\dagger$ \cite{papandreou2015weakly} & 89.2 & 46.7 & 88.5 & 63.5 & 68.4 & 87.0 & 81.2 & 86.3 & 32.6 & 80.7 & 62.4 & 81.0 & 81.3 & 84.3 & 82.1 & 56.2 & 84.6 & 58.3 & 76.2 & 67.2 & 73.9 \\
RNN$^\dagger$ \cite{zheng2015conditional} & 90.4 & 55.3 & 88.7 & 68.4 & 69.8 & 88.3 & 82.4 & 85.1 & 32.6 & 78.5 & 64.4 & 79.6 & 81.9 & \textbf{86.4} & 81.8 & 58.6 & 82.4 & 53.5 & 77.4 & 70.1 & 74.7 \\
BoxSup$^\dagger$ \cite{dai2015boxsup} & 89.8 & 38.0 & \textbf{89.2} & \textbf{68.9} & 68.0 & 89.6 & 83.0 & 87.7 & 34.4 & 83.6 & \textbf{67.1} & 81.5 & 83.7 & 85.2 & 83.5 & 58.6 & 84.9 & 55.8 & \textbf{81.2} & 70.7 & 75.2 \\
Piecewise$^\dagger$ \cite{lin2015efficient} & \textbf{94.1} & 40.7 & 84.1 & 67.8 & \textbf{75.9} & \textbf{93.4} & 84.3 & \textbf{88.4} & \textbf{42.5} & \textbf{86.4} & 64.7 & \textbf{85.4} & \textbf{89.0} & 85.8 & \textbf{86.0} & \textbf{67.5} & \textbf{90.2} & \textbf{63.8} & 80.9 & \textbf{73.0} & \textbf{78.0} \\
\hline\hline
DPN & 87.7 & 59.4 & 78.4 & 64.9 & 70.3 & 89.3 & 83.5 & 86.1 & 31.7 & 79.9 & 62.6 & 81.9 & 80.0 & 83.5 & 82.3 & 60.5 & 83.2 & 53.4 & 77.9 & 65.0 & 74.1 \\
DPN$^\dagger$ & 89.0 & \textbf{61.6} & 87.7 & 66.8 & 74.7 & 91.2 & 84.3 & \textbf{87.6} & 36.5 & \textbf{86.3} & 66.1 & 84.4 & 87.8 & \textbf{85.6} & 85.4 & 63.6 & 87.3 & 61.3 & 79.4 & 66.4 & \textbf{77.5} \\
\hline
\end{tabular}
\subcaption{\footnotesize Per-class results on VOC12 {\em test}. The approaches pre-trained on COCO \cite{lin2014microsoft} are marked with $^\dagger$.}
\end{subtable}
\end{table*}

\begin{table*}
  \scriptsize
	\caption{ Per-class results on Cityscapes.}
	\label{tab:Cityscapes_perclass}
  \centering
  \begin{tabular}{l|p{8pt}p{8pt}p{8pt}|p{8pt}p{8pt}p{8pt}p{8pt}p{8pt}p{8pt}p{8pt}p{8pt}p{8pt}p{8pt}p{8pt}p{8pt}p{8pt}p{8pt}p{8pt}p{8pt}p{8pt}p{8pt}p{8pt}|p{12pt}}
    \hline
    &coarse&depth&sub&road&swalk&build.&wall&fence&pole&tlight&sign&veg.&terrain&sky&person&rider&car&truck&bus&train&mbike&bike&mIoU\\
    \hline\hline
    Dilation10 \cite{yu2015multi} &no&no&no&\textbf{97.6}&\textbf{79.2}&\textbf{89.9}&37.3&47.6&\textbf{53.2}&58.6&65.2&\textbf{91.8}&69.4&93.7&\textbf{78.9}&55&\textbf{93.3}&45.5&53.4&47.7&52.2&66&\textbf{67.1}\\
    Piecewise \cite{lin2015efficient} &no&no&no&97.3&78.5&88.4&44.5&\textbf{48.3}&34.1&55.5&61.7&90.1&\textbf{69.5}&92.2&72.5&52.3&91&54.6&61.6&51.6&\textbf{55}&63.1&66.4\\
    SiCNN+CRF&no&yes&no&96.3&76.8&88.8&40&45.4&50.1&\textbf{63.3}&\textbf{69.6}&90.6&67.1&92.2&77.6&\textbf{55.9}&90.1&39.2&51.3&44.4&54.4&66.1&66.3\\
    FCN \cite{long2014fully} &no&no&no&97.4&78.4&89.2&34.9&44.2&47.4&60.1&65&91.4&69.3&93.9&77.1&51.4&92.6&35.3&48.6&46.5&51.6&\textbf{66.8}&65.3\\
    WSSL \cite{papandreou2015weakly} &yes&no&2&97.4&78.3&88.1&47.5&44.2&29.5&44.4&55.4&89.4&67.3&92.8&71&49.3&91.4&\textbf{55.9}&\textbf{66.6}&\textbf{56.7}&48.1&58.1&64.8\\
    DeepLab \cite{chen2014semantic} &no&no&2&97.3&77.7&87.7&43.6&40.5&29.7&44.5&55.4&89.4&67&92.7&71.2&49.4&91.4&48.7&56.7&49.1&47.9&58.6&63.1\\
    RNN  \cite{zheng2015conditional}  &no&no&2&96.3&73.9&88.2&\textbf{47.6}&41.3&35.2&49.5&59.7&90.6&66.1&93.5&70.4&34.7&90.1&39.2&57.5&55.4&43.9&54.6&62.5\\
    \hline\hline
    DPN&no&no&no&97.5&78.5&89.5&40.4&45.9&51.1&56.8&65.3&91.5&69.4&\textbf{94.5}&77.5&54.2&92.5&44.5&53.4&49.9&52.1&64.8&\textbf{66.8}\\
    \hline
  \end{tabular}
\end{table*}

\begin{table*}
  \scriptsize
	\caption{ Per-class results on CamVid.}
    \label{tab:CamVid_perclass}
  \centering
  \begin{tabular}{l|p{8pt}p{8pt}p{8pt}p{8pt}p{8pt}p{8pt}p{8pt}p{8pt}p{8pt}p{8pt}p{14pt}|p{12pt}}
    \hline
    &build.&tree&sky&car&sign&road&person&fence&pole&pave.&cyclist&mIoU\\
    \hline\hline
    ALE \cite{russell2009associative}&73.4&70.2&91.1&64.24&24.4&91.1&29.1&\textbf{31}&13.6&\textbf{72.4}&28.6&53.59\\
    SuperParsing \cite{tighe2010superparsing}&70.4&54.8&83.5&43.3&25.4&83.4&11.6&18.3&5.2&57.4&8.9&42.03\\
	Tripathi \etal \cite{tripathi2015semantic}&74.2&67.9&91&66.5&23.6&90.7&26.2&28.5&\textbf{16.3}&71.9&28.2&53.18\\
    Liu and He \cite{liu2015multiclass}&66.8&66.6&90.1&62.9&21.4&85.8&28&17.8&8.3&63.5&8.5&47.2\\
    SegNet \cite{badrinarayanan2015segnet}&68.7&52&87&58.5&13.4&86.2&25.3&17.9&16.0&60.5&24.8&46.4\\
    \hline\hline
    DPN&\textbf{80.6}&72.6&91.2&77.8&\textbf{40}&90.7&\textbf{43.9}&28.7&15.9&71.4&\textbf{47.9}&60.06\\
    Spatial-temporal DPN&\textbf{80.6}&\textbf{73.1}&\textbf{91.4}&\textbf{77.9}&\textbf{40}&\textbf{90.8}&\textbf{43.9}&29.2&16&71.9&\textbf{47.9}&\textbf{60.25}\\
    \hline
  \end{tabular}
 
\end{table*}

\vspace{4pt}\hspace{-15pt}\textbf{Per-stage Analysis.~}
We further evaluate DPN using three metrics.
The results are given in Fig.~\ref{fig:error}.
For example, (a) illustrates that the tagging accuracy can be improved in the third stage, as it captures label co-occurrence with a mixture of contextual patterns. However, TA decreases a little after the final stage.
%
Since joint tuning maximizes segmentation accuracies by optimizing all components together,
extremely small objects, which rarely occur in VOC training set, are discarded.
%
%
%
As shown in (b), accuracies of object localization are significantly improved in the second and the final stages.
This is intuitive because the unary prediction can be refined by long-range and high-order pixel relations, and joint training further improves results.
(c) discloses that the second stage also captures object boundary, since it measures dissimilarities between pixels.

\vspace{4pt}\hspace{-15pt}\textbf{Per-class Analysis.~}
Table \ref{tab:perclass}(a) reports the per-class accuracies of four evaluation metrics, where the first four rows represent the mIoU of four stages, while the last three rows represent TA, LA, and BA, respectively.
We have several valuable observations, which motivate future researches.
(1) Joint training benefits most of the categories, except animals such as `bird', `cat', and `cow'.
Some instances of these categories are extremely small so that joint training discards them for smoother results.
(2) Training DPN with pixelwise label maps implicitly models image-level tags, since it achieves a high averaged TA of 96.4\%.
(3) Object localization always helps.
However, for objects with complex boundary such as `bike', its mIoU is low even it can be localized, \eg `bike' has high LA but low BA and mIoU.
%
%
(4) Failures of different categories have different factors.
With these three metrics, they can be easily identified.
%
For example, the failures of  `chair', `table', and `plant'  are caused by the difficulties to accurately capture their bounding boxes and boundaries.
%
Although `bottle' and `tv' are also difficult to localize, they achieve moderate mIoU because of their regular shapes.
In other words, mIoU of `bottle' and `tv' can be significantly improved if they can be accurately localized.

\subsection{Benchmarks}\label{sec:overall}

We evaluate the performance of DPN and spatio-temporal DPN on several standard semantic segmentation benchmarks.

\subsubsection{Pascal VOC12}
%
%
The VOC12 dataset is one of the most popular benchmarks for semantic image segmentation. This dataset contains 20 indoor and outdoor object categories and one background category. As previously mentioned, we employ $10,582$ images for training, $1,449$ images for validation, and $1,456$ images for testing. Results are given Table \ref{tab:perclass} (b), we compare DPN with the best-performing methods\footnote{The results of these methods were presented in either the published papers or arXiv pre-prints.} on VOC12 test set based on two settings, \ie with and without pre-training on COCO.
%
The approaches pre-trained on COCO are marked with `$\dag$'.
We evaluate DPN on several scales of the images and then average the results following \cite{chen2014semantic, lin2015efficient}.
%

DPN outperforms several existing methods that were trained on VOC12, but DPN needs only one MF iteration to solve MRF, other than 10 iterations of RNN, DeepLab, and Piecewise.
By averaging the results of two DPNs,
we achieve 74.1\% accuracy on VOC12 without outside training data.
%
As discussed in Sec.\ref{sec:learn}, MF iteration is the most complex step even when it is implemented as convolutions.
Therefore, DPN at least reduces 10$\times$ runtime compared to previous works.

Following \cite{zheng2015conditional, dai2015boxsup}, we pre-train DPN with COCO, where 20 object categories that are also presented in VOC12 are selected for training.
%
%
%
A single DPN$^\dagger$ has achieved 77.5\% mIoU on VOC12 test set.
%
%
%
{The system in~\cite{lin2015efficient} has incorporated multi-scale pyramid training and mid-level feature refinement,
which we believe is complementary to our contribution and can boost the performance of DPN as well.
As shown in Table \ref{tab:perclass} (b), we observe that DPN$^\dagger$ achieves competitive performances among multiple object classes.}

\subsubsection{Cityscapes} 

Cityscapes dataset focuses on street scenes segmentation. 
All images are captured from a moving vehicle in various seasons and cities. 
This dataset is highly different from VOC12. 
Because of the great depth of images, the street scenes have large scale variations even in a single image. 
It defines 19 object categories for evaluation. 
There are 2975 training, 500 validation and 1525 testing images with fine pixel-level annotations and extra 20000 images with cores annotations. 
To conduct a fair comparison, we only use the images with fine pixel-level annotations in our experiment.

As shown in Table \ref{tab:Cityscapes_perclass}, DPN achieves 66.8\% on Cityscapes dataset, which is the second best method, and it is close to the first place 67.1\% \cite{yu2015multi}. 
\cite{yu2015multi} takes a multi-scale aggregation strategy, which can be easily integrated into DPN to further boost the performance.
Our method shows advantages on objects with arbitrary shape (\ie `road', `building', `vegetation, `terrain' and `sky') which can be easily confused with other objects. Thanks to our high-order and long-range pairwise term, DPN captures their appearance and segments them from other objects accurately. 

\begin{figure*}[t]
  \centering
  \includegraphics[width=0.72\textwidth]{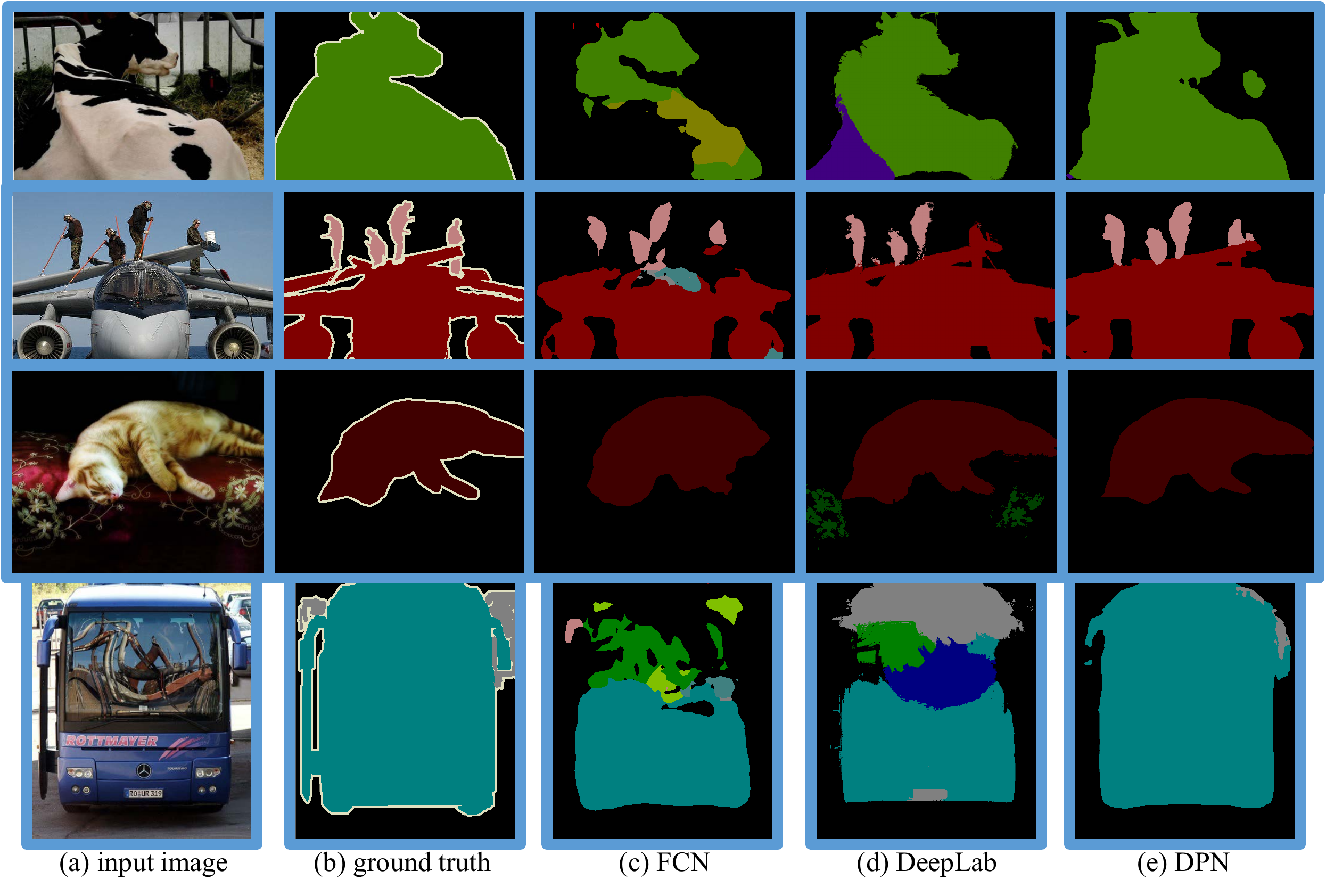}
  \caption{Visual quality comparison of different semantic image segmentation methods: (a) input image (b) ground truth (c) FCN \cite{long2014fully} (d) DeepLab \cite{chen2014semantic} and (e) DPN.}
  \label{fig:visualquality_comparison}
\end{figure*}

\subsubsection{CamVid}

CamVid dataset consists of 367 training and 233 testing images with 11 classes annotations. 
All images are extracted from three video sequences at 1Hz. 
Similar to Cityscapes dataset, these video sequences are also captured from a moving vehicle. 
As shown in Table \ref{tab:CamVid_perclass},  DPN outperforms all existing methods, and spatial-temporal pairwise term further improves the performance to 60.25\%.
Considering the relatively sparse sampling rate (one frame per second) and strong unary term used, our 3-D MRF can indeed leverage temporal contextual information for joint inference.  

We can observe that DPN achieves much better performance than other methods, especially on narrow and small objects (\ie `pole' and `sign'), which are very difficult in segmentation task. Since our triple penalty captures the appearance of pixels, we can predict more accurate boundary on those objects. 
The spatial-temporal pairwise term further improves the performance. 
Spatial-temporal DPN encodes the relationship between successive frames, which improve the performance on some specific categories like `tree', `sky' and `road', because they have flexible shape and are continuous in successive frames. 
It is also worthwhile to note that our spatial-temporal DPN achieves best performances on 7 object categories out of 11.

\begin{figure*}[t]
	\centering
	\includegraphics[width=0.6\textwidth]{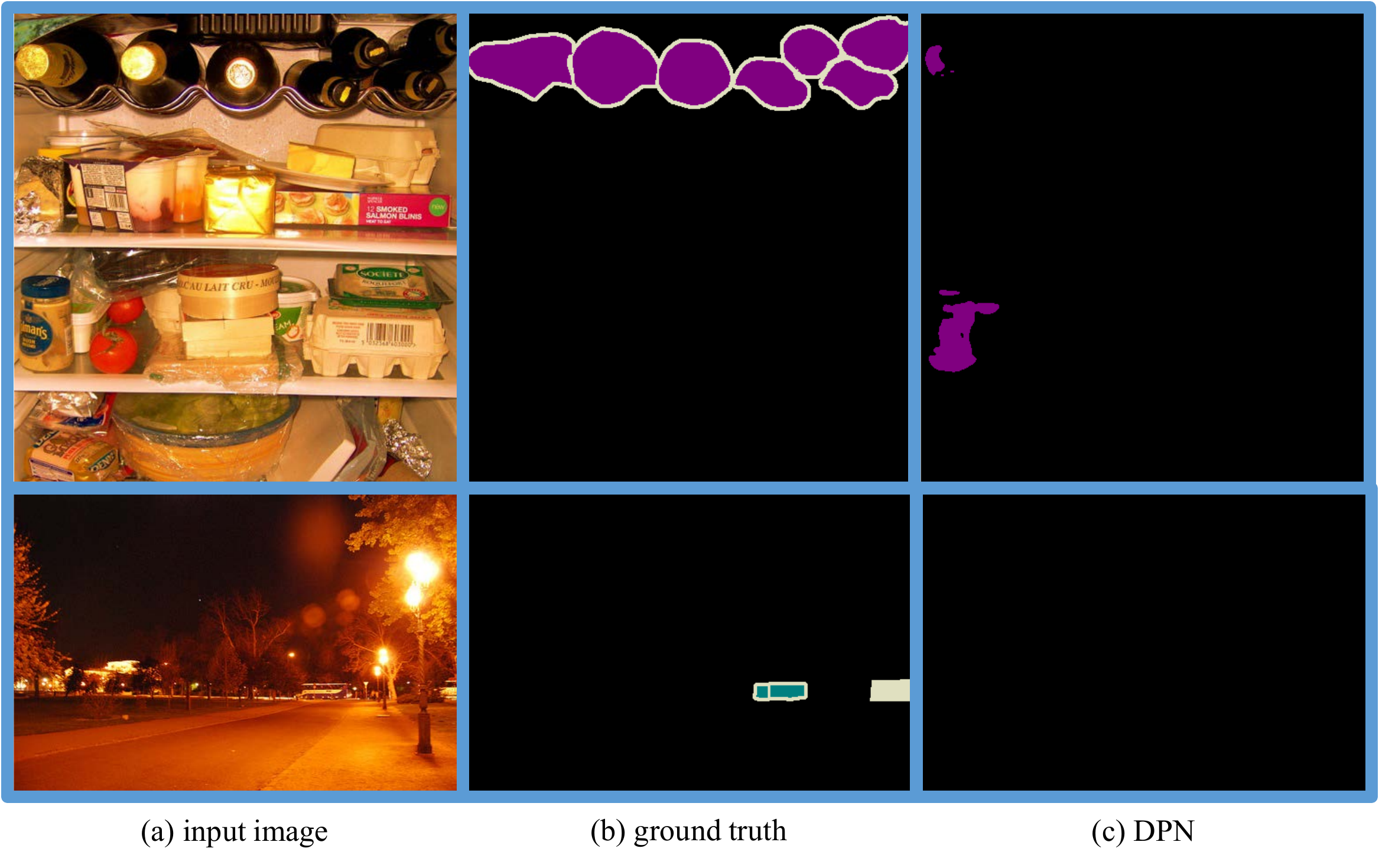}
	\caption{Failure cases: (a) input image (b) ground truth (c) DPN.}
	\label{fig:failure_case}
\end{figure*}

\begin{figure*}
  \centering
  \includegraphics[width=0.8\textwidth]{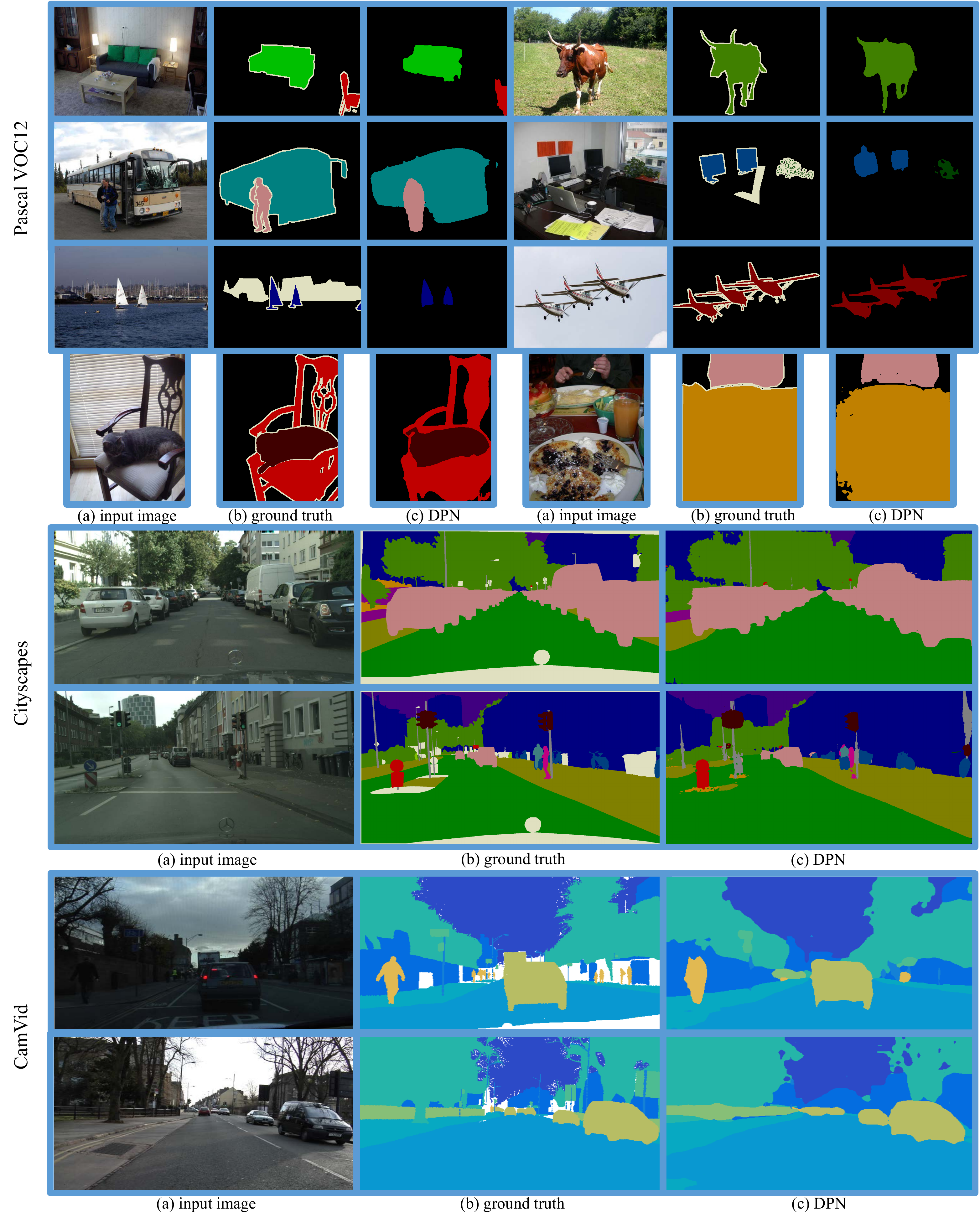}
  \caption{Visual quality of DPN label maps: (a) input image (b) ground truth (white labels indicating ambiguous regions) and (c) DPN.}
  \label{fig:visualquality_self}
\end{figure*}

\subsection{Visual Quality Comparisons}

In the following, we inspect visual quality of obtained label maps.
Fig.~\ref{fig:visualquality_comparison} demonstrates the comparisons of DPN with FCN \cite{long2014fully} and DeepLab \cite{chen2014semantic}.
We use the publicly released model\footnote{\url{http://dl.caffe.berkeleyvision.org/fcn-8s-pascal.caffemodel}} to re-generate label maps of FCN while the results of DeepLab are extracted from their published paper.
DPN generally makes more accurate predictions in both image-level and instance-level.
For example, in Fig.~\ref{fig:visualquality_comparison} (row 2), DPN is able to discover all persons available while still recovering sharp boundaries of the aeroplane.
Even in the challenging case of Fig.~\ref{fig:visualquality_comparison} (row 4) due to complex reflectance, the integrity of the bus is maintained by DPN.
More examples of DPN label maps are shown in Fig.~\ref{fig:visualquality_self}.
We observe that learning local label contexts helps differentiate confusing objects and learning triple penalty facilitates the capturing of intrinsic object boundaries.

We also include some typical failure modes of DPN in Fig.~\ref{fig:failure_case}. 
In the first case, the object with atypical pose is hard to be found. 
Further considering well-trained object detector as additional unary potential might be a potential solution to this challenge.  
The second case suggests that scale and illumination are also important factors that influence the performance. 
This problem can be partially alleviated by augmenting or adding the variance of training data.

\section{Conclusion}

We proposed Deep Parsing Network (DPN) to address semantic image/video segmentation. DPN has several appealing properties.
First, DPN unifies the inference and learning of unary term and pairwise terms in a single convolutional network.
No iterative inference is required during back-propagation.
Second, high-order relations and mixtures of label contexts are incorporated to its pairwise terms modeling,
making existing works as special cases.
Third, DPN is built upon conventional operations of CNN, thus easy to be parallelized and speeded up.

DPN achieves state-of-the-art performance on VOC12, Cityscapes and CamVid datasets.
Multiple valuable facts about semantic segmentation are revealed through extensive experiments, such as the interplay between tagging, localization and boundary accuracies along different processing stages.
%
Future directions include investigating the generalizability of DPN to more challenging scenarios with large number of object classes and substantial appearance/scale variations.

\section*{Acknowledgment}

This work is supported by SenseTime Group Limited and the General Research Fund sponsored by the Research Grants Council of the Hong Kong SAR (CUHK 416713, 14241716, 14224316. 14209217).


%
%

\ifCLASSOPTIONcaptionsoff
  \newpage
\fi


%
%
\bibliographystyle{IEEEtran_zwliu}
\bibliography{egbib,natbib}

%
\begin{IEEEbiography}[{\includegraphics[width=1in,height=1.25in,clip,keepaspectratio]{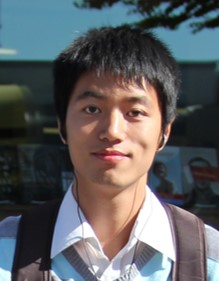}}]{Ziwei Liu}
received his B.E. degree in Department of Electronics and Information Engineering in 2013, from Huazhong University of Science and Technology, Wuhan, China. He is currently a PhD student at the Department of Information Engineering, The Chinese University of Hong Kong. His research interests include computer vision, machine learning and computational photography, with focus on human-centric visual understanding and deep learning.
\end{IEEEbiography}

\begin{IEEEbiography}[{\includegraphics[width=1in,height=1.25in,clip,keepaspectratio]{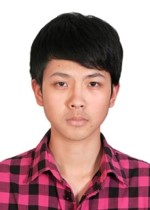}}]{Xiaoxiao Li}
received his B.E. degree in Department of Computer Science and Technology from Tsinghua University in 2014. He is currently a PhD student at the Department of Information Engineering, The Chinese University of Hong Kong. His research interests include computer vision and machine learning, especially image segmentation and deep learning. 
\end{IEEEbiography}

\begin{IEEEbiography}[{\includegraphics[width=1in,height=1.25in,clip,keepaspectratio]{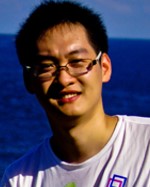}}]{Ping Luo} received his PhD degree in 2014 in Information Engineering, Chinese University of Hong Kong (CUHK). He is currently a Research Assistant Professor in Electronic Engineering, CUHK. His research interests focus on deep learning and computer vision, including optimization, face recognition, web-scale image and video understanding. He has published 40+ peer-reviewed articles in top-tier conferences such as CVPR, ICML, NIPS and journals such as TPAMI and IJCV. He received a number of awards for his academic contribution, such as Microsoft Research Fellow Award in 2013 and Hong Kong PhD Fellow in 2011. He is a member of IEEE. 
\end{IEEEbiography}

\begin{IEEEbiography}[{\includegraphics[width=1in,height=1.25in,clip,keepaspectratio]{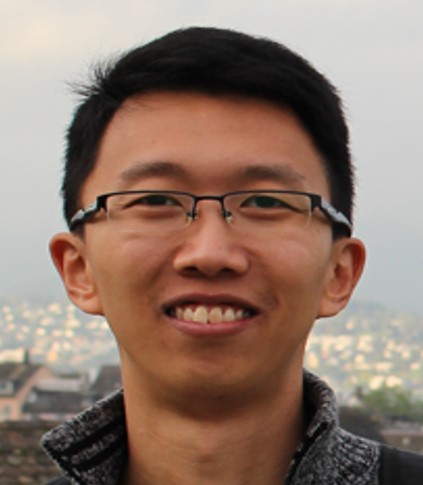}}]{Chen Change Loy} (S'06-M'10-SM'17) received the PhD degree in computer science from the Queen Mary University of London in 2010. He is currently a research assistant professor in the Department of Information Engineering, Chinese University of Hong Kong. Previously, he was a postdoctoral researcher at Queen Mary University of London and Vision Semantics Ltd. His research interests include computer vision and pattern recognition, with focus on face analysis, deep learning, and visual surveillance. He serves as an Associate Editor of IET Computer Vision Journal. He is a senior member of the IEEE.
\end{IEEEbiography}

\begin{IEEEbiography}[{\includegraphics[width=1in,height=1.25in,clip,keepaspectratio]{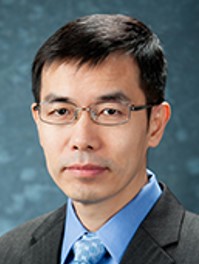}}]{Xiaoou Tang} (S'93-M'96-SM'02-F'09) received the B.S. degree from the University of Science and Technology of China, Hefei, in 1990, and the M.S. degree from the University of Rochester, Rochester, NY, in 1991. He received the Ph.D. degree from the Massachusetts Institute of Technology, Cambridge, in 1996. He is a Professor and the Chairman of the Department of Information Engineering, Chinese University of Hong Kong. He worked as the group manager of the Visual Computing Group at the Microsoft Research Asia from 2005 to 2008. His research interests include computer vision, pattern recognition, and video processing. He received the Best Paper Award at the IEEE Conference on Computer Vision and Pattern Recognition (CVPR) 2009 and Outstanding Student Paper Award at the AAAI 2015. He was a program chair of the IEEE International Conference on Computer Vision (ICCV) 2009 and served as an associate editor of the IEEE Transactions on Pattern Analysis and Machine Intelligence and International Journal of Computer Vision. He is now the Editor-in-Chief of the International Journal of Computer Vision. He is a fellow of the IEEE.
\end{IEEEbiography}






\end{document}